# A Comprehensive Review of Deep Learning-based Single Image Super-resolution

**Syed Muhammad Arsalan Bashir[1, 2,*]*, Yi Wang[1], and Mahrukh Khan[3]**

[1] School of Electronics and Information, Northwestern Polytechnical University, Xi'an, Shaanxi, China
[2] Space and Upper Atmosphere Research Commission, Karachi, Sindh, Pakistan
[3] Department of Computer Science, National University of Computer and Emerging Sciences, Karachi, Sindh, Pakistan

## ABSTRACT

Image super-resolution (SR) is one of the vital image processing methods that improve the resolution of an image in the field of computer vision. In the last two decades, significant progress has been made in the field of super-resolution, especially utilizing deep learning methods. This survey is an effort to provide a detailed survey of recent progress in the field of super-resolution in the perspective of deep learning while also informing about the initial classical methods used for achieving super-resolution. The survey classifies the image SR methods into four categories, i.e., classical methods, supervised learning-based methods, unsupervised learning-based methods, and domain-specific SR methods. We also introduce the problem of SR to provide the intuition about image quality metrics, available reference datasets, and SR challenges. Deep learning-based approaches of SR are evaluated using a reference dataset. Finally, this survey is concluded with future directions and trends in the field of SR and open problems in SR to be addressed by the researchers.

[*] Corresponding author: smarsalan@mail.nwpu.edu.cn; smab1176@yahoo.com; ORCID: 0000-0002-9899-6293

# 1 INTRODUCTION

The image-based computer graphics models lack resolution independence [1] as the images cannot be zoomed beyond the image sample resolution without compromising the quality of images. Thus, simple image interpolation will lead to blurring of features and edges within a sample image.

The concept of super-resolution was first used by [2] to improve the resolution of an optical system beyond the diffraction limit. In the past two decades, the concept of super-resolution (SR) is defined as the method of producing high-resolution (HR) images from a corresponding low-resolution (LR) image. Initially, this technique was classified as spatial resolution enhancement [3]. The applications of super-resolution include computer graphics [4–6], medical imaging [7–12], security, and surveillance [13,14], which shows the importance of this topic in recent years.

The image super-resolution, although being explored for decades, remains a challenging task in computer vision, and this problem is fundamentally ill-posed because, for any given LR image, there can be several HR images with slight variations in camera angle, color, brightness, and other variables. Furthermore, there are fundamental uncertainties among the LR and HR data since the downsampling of different HR images may lead to a similar LR image, which makes this conversion a many-to-one process [15].

In the past, classical SR methods such as statistical methods, prediction-based methods, patch-based methods, edge-based, and sparse representation methods were used to achieve super-resolution. However, recently the advances in computational power and big data have made researchers use deep learning (DL) to address the problem of SR. In the past decade, deep learning-based SR studies have reported superior performance than the classical methods, and DL methods have been used frequently to achieve SR. A range of methods has been used by researchers to explore SR, ranging from the first method of Convolutional Neural Network (CNN) [16] to the recently used Generative Adversarial Nets (GAN) [17]. In principle, the methods used in deep learning-based SR methods vary in hyper-parameters such as network architecture, learning strategies, activation functions, and loss functions.

In this study, a brief overview of the classical methods of SR is outlined initially, whereas the main focus is given to give an overview of the most recent researches in the field of SR using deep learning. Previous studies have explored the literature on SR, but most of these studies emphasize the classical methods [18–22], additionally [21,22] used human visual perception to gauge the performance of SR methods. This survey is a comprehensive overview of the recent advances in SR with an emphasis on deep learning-based approaches and their achievements in systematically achieving SR. Tables 3 and 4 in the supplementary materials depict the full list of symbols and acronyms used in this study.

In the next sections, the problem definition, along with the associated concepts of image super-resolution, are discussed in light of the literature review.

## 1.1 SUPER-RESOLUTION - DEFINITION

The image SR focuses on the recovery of an HR image from LR image input as and in principle, the LR image $I_{xLR}$ can be represented as the output of the degradation function, as shown in (1).

$$I_{xLR} = d\left(I_{yHR}, \partial\right) \tag{1}$$

Where d is the SR degradation function that is responsible for the conversion of HR image to LR image, $I_{yHR}$ is the input HR image (reference image), whereas $\partial$ depicts the input parameters of the image degradation function. Degradation parameters are usually scaling factor, blur type, and noise. In practice, the degradation process and dependent parameters are unknown, and only LR images are used to get HR images by SR method. The SR process is responsible for predicting the inverse of the degradation function d, such that $g = d^{-1}$

$$g\left(I_{xLR}, \delta\right) = d^{-1}\left(I_{xHR}\right) = I_{yE} \approx I_{yHR} \tag{2}$$

Where g is the SR function, $\delta$ depicts the input parameters to the function $g$, and $I_{yE}$ is the estimated HR corresponding to the input $I_{xLR}$ image. It is also worth noticing the super-resolution function, as in (2), is ill-posed, as the function $g$ is a non-injective function; thus, there are infinite possibilities of $I_{yE}$ for which the condition $d\left(I_{yE}, \partial\right) = I_{xLR}$ will hold.

The degradation process for the input LR images is unknown, and this process is affected by numerous factors such as sensor-induced noise, artifacts created because of lossy compression, speckle noise, motion blur, and misfocused images. In the literature, most of the studies have used a single downsampling function as the image degradation function:

$$d\left(I_{yHR}, \partial\right) = \left(I_{yHR}\right)\downarrow_{s_f}, \{s\} \subseteq \partial \tag{3}$$

Where $\downarrow s_f$ is the downsampling operator with $s_f$ being the scaling factor. One of the frequently used downsampling functions in SR is the bicubic interpolation [23–25] with antialiasing. In some studies, like [26], researchers have used more operations in the downsampling function, and the overall downsampling operation is:

$$d\left(I_{yHR},\partial\right)=\left(I_{yHR}\otimes\kappa\right)\downarrow_{s_f}+n_{\sigma},\{\kappa,s,\sigma\}\subseteq\partial \tag{4}$$

Where $I_{yHR}\otimes\kappa$ depicts the convolution of the HR image $I_{yHR}$ with the blurring kernel $\kappa$, $n_{\sigma}$ represents the additive white Gaussian noise with a standard deviation of $\sigma$. The degradation function defined in (4) and Fig. 1 is closer to the actual function as it considers more parameters than the simple downsampling degradation function [26].\

Finally, the purpose of SR is to minimize the loss function as follows:

$$\hat{\phi}=\left[\min\mathcal{L}\left(I_{yE},I_{yHR}\right)\right]_{\phi}+h\Psi(\phi) \tag{5}$$

Where $\mathcal{L}(I_{yE},I_{yHR})$ is the loss function between output HR image of SR and the actual HR image, h is the tradeoff parameter whereas $\Psi(\phi)$ is the regularization term. The most common loss function used in SR is the pixel-based mean square error (MSE), which can also be referred to as the pixel loss. In recent years, researchers have used a combination of various loss functions, and these combinations are further explored in later sections. Further mathematical modeling of the SR problem is discussed in [27].

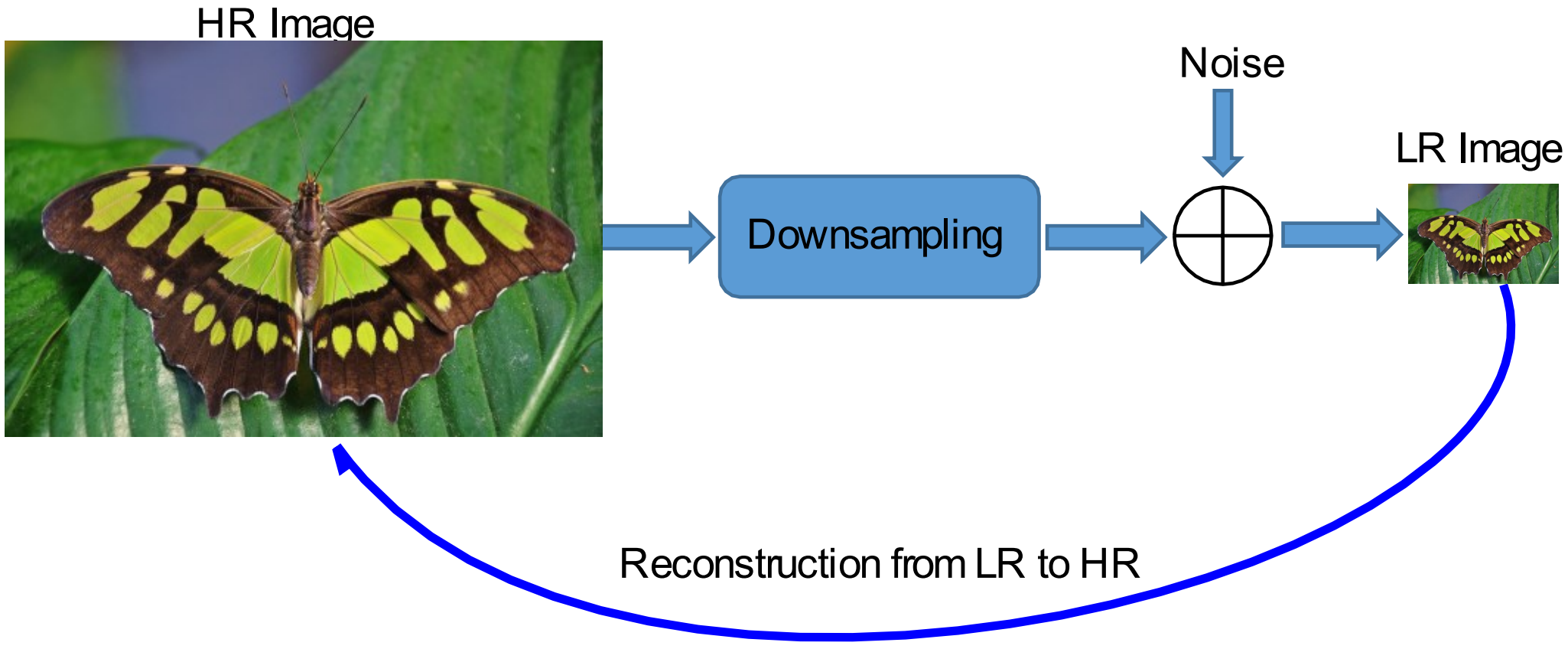

Fig. 1. Downsampling and upsampling in super-resolution, noise is added to simulate realistic degradation within an image.

## 1.2 METHODS FOR QUALITY OF SR IMAGES

Image quality can have several definitions per measurement methods, and it is generally a measure of the quality of visual attributes and perception of the viewers. The image quality assessment (IQA) methods are characterized into subjective methods (human perception of an image is natural and of good quality) and objective methods (quantitative methods by which image quality can be numerically computed) [28].

Quality related visual aspects of an image are mostly a good measure, but this method requires more resources, especially if the dataset is large [29]; thus, in SR and computer vision tasks, the more suitable methods are objective. As per [30], the IQA methods are primarily categorized into three categories, i.e., reference image-based features from the actual image, and blind IQA with no information about the ground truth. In this section, IQA methods primarily used in the domain of SR are further explored.

### 1.2.1 PEAK SIGNAL-TO-NOISE RATIO

As in most of the information systems, the peak signal-to-noise ratio (PSNR) is a measurement technique for analyzing the signal power in comparison to the power of noise, especially in images, the PSNR is used as a quantitative measure of the compression quality of an image. In the field of super-resolution, the PSNR of an image is defined by the maximum pixel value and the mean square error between the reference image in comparison to the SR image, also known as the power of image distortion noise. For a given maximum pixel value ($M$) and the reference image ($I_r$) having t pixels and the SR image ($I_y$), the peak signal-to-noise ratio is defined as:

$$PSNR=10\log_{10}\left(\frac{M^2}{MSE}\right) \tag{6}$$

Where $M$ is mostly for 8-bit color space depth, i.e., the max value of 255 and MSE is given by:

$$MSE=\frac{1}{t}\sum_{i=1}^{t}\left(I_r(i)-I_y(i)\right)^2 \tag{7}$$

As seen from (6), the PSNR is related to the individual pixel intensity values of the SR image and reference image and is a pixel-based metric of image quality. In some cases [31–33], this quality metric can be misleading as the overall image might not be visually similar to that of the reference image. This metric is still used for image comparisons, especially comparing the results of SR algorithms with previously published results to compare the working of any new method in the field of SR.

MSE for color images averaged for color channels and an alternate approach is to measure PSNR for luminance and or greyscale channels separately as the human eye is more sensitive to changes in luminance in contrast to changes in chrominance [34].

### 1.2.2 STRUCTURAL SIMILARITY INDEX

The visual perception of humans, is efficient in extracting the structural information within an image, and PSNR does not consider the structural composition of the image [35]. The structural similarity index metric (SSIM) was proposed by [36] to measure the structural similarity between images by comparing the contrast, luminance, and structural details within the reference image.

An image $I_r$ with total pixels $P$; the contrast $C_I$, and luminance $L_I$ can be denoted as the standard deviation and the mean of the image intensity given by:

$$L_I = \frac{1}{M}\sum_{i=1}^{M} I_r(i) \tag{8}$$

$$C_I = \sqrt{\left(\frac{1}{M-1}\sum_{i=1}^{M}\left(I_r(i)-L_I\right)^2\right)} \tag{9}$$

The ith pixel of the reference image is denoted by $I_r(i)$. The comparisons based on the contrast and luminance between the reference image and the estimated image are:

$$Com_l\left(I_r, \hat{I}\right) = \frac{2L_{I_r}L_{\hat{I}} + \mu_1}{L_{I_r}^2 + L_{\hat{I}}^2 + \mu_1} \tag{10}$$

$$Com_c\left(I_r, \hat{I}\right) = \frac{2C_{I_r}C_{\hat{I}} + \mu_2}{C_{I_r}^2 + C_{\hat{I}}^2 + \mu_2} \tag{11}$$

Where $\mu_1 = (k_1 S)^2$ and $\mu_2 = (k_2 S)^2$, these constant terms ensure stability by ensuring $k_1 << 1$ and $k_2 << 1$.

Normalized pixel values $I_r - L_{I_r}/C_{I_r}$ represent the image structure, while the inner product of these is the equivalent of structural similarity between the reference image $I_r$ and estimated image $\hat{I}$. The covariance is given by:

$$\sigma_{I_r,\hat{I}} = \frac{1}{M-1}\sum_{i=1}^{M}\left(I_r(i) - L_{I_r}\right)\left(\hat{I}(i) - L_{\hat{I}}\right) \tag{12}$$

Function for structural comparison $Com_s\left(I_r, \hat{I}\right)$ is given by:

$$Com_s\left(I_r, \hat{I}\right) = \frac{\sigma_{I_r,\hat{I}} + \mu_3}{C_{I_r}C_{\hat{I}} + \mu_3} \tag{13}$$

Where $\mu_3$ is stability constant, the final structural similarity index (SSIM) is given by:

$$SSIM\left(I_r, \hat{I}\right) = \left\{Com_l\left(I_r, \hat{I}\right)\right\}^{\alpha}\left\{Com_c\left(I_r, \hat{I}\right)\right\}^{\beta}\left\{Com_s\left(I_r, \hat{I}\right)\right\}^{\gamma} \tag{14}$$

The control parameters are $\alpha$, $\beta$, and $\gamma$, which can be adjusted to increase the importance of luminance, contrast, and structural comparison in calculating the SSIM.

Conventionally, PSNR is used in computer vision tasks for evaluation, but SSIM is based on human perception of structural information within an image, thus this method is widely used for comparing the structural similarity between images [37,38].

### 1.2.3 OPINION SCORING

Opinion scoring is a qualitative method, which lies in the subjective category of IQA. In this method, the quality testers are asked to grade the quality of images based on specific criteria, e.g., sharpness, natural look, and color, where the final graded score is the mean of the rated scores.

This method has limitations such as non-linearity between the scores, variation in results due to changes in test criteria, and human error. In SR, certain methods have reported good objective quality scores but scored poorly in subjective results, especially in the case of human face reconstruction [17], [39,40]. Thus, the opinion scoring method is also used in studies [29], [41–44] to measure quality of human perception

### 1.2.4 PERCEPTUAL QUALITY

Opinion scoring used human raters for manual evaluation of the images, while this method can provide accurate results as far as human perception is concerned; this method requires many resources, especially for large datasets [45]. Initially, [46] proposed a CNN based full reference image quality assessment (FR-IQA) model where human behavior was learned using an IQA database that contained distorted images, subjective scores, and error maps, and this method was called DeepQA.

In [47], the authors used quality-discriminable image pairs (DIP) for training and the system was called dipQA (DIP inferred quality index), they used RankNet with L2R algorithm to learn blind opinion IQA whereas in [48] a multi-task end-to-end optimized deep neural network (MEON) was proposed. MEON used two stages, in first stage distortion type learning using large datasets already available, and in the second stage, the output of the first stage was used to train the quality assessment network using stochastic gradient descent. In [49], the authors used CNN to develop a no-reference IQA method known as NIMA; NIMA was trained on pixel-level and aesthetic quality datasets.

RankIQA [50] trained a Siamese network to grade the quality of images using datasets with known image distortions, CNNs were used to learn the IQA, and this method even outperformed full-reference methods without using the reference image. IQA proposed in [51] included ten convolution layers along with five pooling layers for feature extraction while there were two fully connected layers for regression, this method performed significantly well for both no-reference and full-reference IQA.

Even though opinion scoring and perceptual quality-based methods, do exhibit human perception in IQA, but the quality we require is still an open question (i.e., if we want images to be more natural or similar to the reference image) thus PSNR and SSIM are the primarily used methods in computer vision and SR

### 1.2.5 TASK-BASED EVALUATION

Although the primary purpose of image SR is to achieve better resolution as mentioned earlier, SR is also helpful in other computer vision tasks [5], [6], [12], [52]. The performance achieved in these can be used to indirectly measure the performance of the SR methods used in those tasks. In the case of medical images, the original and SR constructed images were used by the researchers to see the performance in the training and prediction phases. In general, computer vision tasks such as classification [53], [54], face recognition [39,40], [55], and object segmentation [56–58] can be done using SR images, and their performance can be used as a metric to assess the performance of the SR method.

### 1.2.6 MISCELLANEOUS IQA METHODS

Development of IQA methods is an open field, and in recent years various researchers have proposed SR metrics, but these methods did were not used widely by the SR community. Feature similarity (FSIM) index metric [59] evaluates image quality by extracting feature points that are considered by the human visual system based on gradient magnitude and phase congruency. The multi-scale structural similarity (MS-SSIM) [60] used multi-scale to incorporate variations in the viewing conditions to measure the image quality and proposed that MS-SSIM provides form flexibility in the measurement of image quality than single-scale SSIM. In [61], the authors claimed that SSIM and MS-SSIM do not perform well on distorted and noisy images, thus they used four-component-based weighted method that adjusted the weight of scores based on the local feature whereas in the case of contrast-distorted images [62] like TID2013 and CSIQ datasets SSIM does not perform well.

According to [37], the perceptual quality and image distortion are at odds with each other; as the distortion decreases, the perceptual quality should also be worse; thus, the accurate measurement of SR image quality is still an open area of research.

## 1.3 OPERATING COLOR CHANNELS

In most of the datasets, RGB color space is used; thus, SR methods mostly employ the use of RGB images, YCbCr space is also used in SR [63]. The Y component in YCbCr is the luminance component, which represents the light intensity, while Cb and Cr are the chrominance components (i.e., blue-differenced and red-differenced Chroma channels) [64]. In recent years, most of the SR challenges and datasets use the RGB color space, which limits the use of RGB space for comparison with state of the art. Furthermore, the results of IQA based on PSNR vary if the color space in the testing stage is different from the training/evaluation stage.

TABLE 1
LIST OF BENCHMARK DATASETS USED IN SR

| Name | Number of images/ pairs | Image format | Type | Resolution | Details of images |
|---|---|---|---|---|---|
| BSD100[58] | 100 | PNG | Unpaired | (480, 320) | 100 images of animals, people, buildings, scenic views etc. |
| BSDS300[58] | 300 | JPG | Unpaired | (430, 370) | 300 images of animals, people, buildings, scenic views, plants, etc. |
| BSDS500[65] | 500 | JPG | Unpaired | (430, 370) | Extended version of BSD 300 with additional 200 images |
| CelebA[55] | 202,599 | PNG | Unpaired | (2048, 1024) | Over 40 attribute defined categories of celebrities |
| DIV2K[66] | 1000 | PNG | Paired | (2048, 1024) | Objects, People, Animals, scenery, nature |
| Manga109[67] | 109 | PNG | Unpaired | (800, 1150) | 109 manga volumes drawn by professional manga artists in Japan |
| MS-COCO[56] | 164,000 | JPG | Unpaired | (640, 480) | Labelled objects with over 80 object categories |
| OutdoorScene[57] | 10624 | PNG | Unpaired | (550, 450) | Outdoor scenes including plants, animals, sceneries, water reservoirs, etc. |
| PIRM[68] | 200 | PNG | Unpaired | (600, 500) | Sceneries, people, flowers, etc. |
| Set14[69] | 14 | PNG | Unpaired | (500, 450) | Faces, animals, flowers, animated characters, insects, etc. |
| Set5[70] | 5 | PNG | Unpaired | (300, 340) | Only 5 images including, butterfly, baby, bird, head, and women. |
| T91[71] | 91 | PNG | Unpaired | (250, 200) | 91 images of fruits, cars, face, etc. |
| Urban100[72] | 100 | PNG | Unpaired | (1000, 800) | Urban buildings, architecture |
| VOC2012[73] | 11,530 | JPG | Unpaired | (500, 400) | Labelled objects with over 20 classes |

## 1.4 DETAILS OF THE REFERENCE DATASET

The datasets used in the evaluation of the SR algorithms are summarized in this section; the various datasets discussed in this section vary in the total number of example images, image resolution, quality, and imaging hardware setup. A few of the datasets comprise paired LR-HR images for training and testing SR algorithms. In contrast, the rest of the datasets include HR images, and the corresponding LR images are usually generated by using bicubic interpolation with antialiasing as performed in [23–25]. Matlab function *imresize (I, scale)*, where the default method is bicubic interpolation with antialiasing and scale is the downsampling factor input of the function.

Table 1 comprises of a list of datasets that are frequently used in SR along with information on total image count, image format, pixel count, HR resolution, type of dataset, and classes of images.

Most of the datasets for SR are unpaired data, and the LR images are generated using various scale factors using bicubic interpolation with antialiasing. Other than the mentioned datasets in Table 1, datasets like General-100 [74], L20 [75], and ImageNet [76] are also used in computer vision tasks. In recent times, researchers have preferred the use of multiple datasets for training/evaluation and testing the SR models; for instance, in [77–80], the researchers used SET5, SET14, BSDS100, and URBAN100 for training and testing.

## 1.5 SUPER-RESOLUTION CHALLENGES

The most prominent SR challenges NTIRE [66], [81], and PIRM [68] are discussed in this section.

The New Trends in Image Restoration and enhancement (NTIRE) challenge [66], [81] was in collaboration with the Conference on Computer Vision and Pattern recognition (CVPR), NTIRE includes various challenges like colorization, image denoising, and SR. In the case of SR, the DIV2K dataset [66] was used, which included bicubic downscaled image pairs and blind images with realistic but unknown degradation. This dataset has been widely used to evaluate SR methods under known and unknown conditions to compare against the state-of-the-art methods.

The perceptual image restoration and manipulation (PRIM) challenges were in collaboration with the European Conference on Computer vision (ECCV), and like NTIRE, it contained multiple challenges. Apart from the three challenges, as mentioned in NTIRE, PRIM also focused on SR for smartphones and the comparison of perceptual quality with the generation accuracy [68]. As mentioned by [37], the models that focus on distortion often give visually unpleasing SR images, while the models focusing on the perceptual image quality do not perform well on information fidelity. Using the image quality metrics NIQE [82] and [83], the methods that performed best in achieving perceptual quality [37] was the winner. In contrast, in a sub-challenge [84], SR methods were evaluated using limited resources to evaluate SR performance for smartphones using the PSNR, MS-SSIM, and opinion scoring metrics. Thus, PIRM encouraged the researchers to explore the perception-distortion tradeoff domain as well as the use of SR for smartphones.

The key features of this study are:

1. We highlight a brief overview of the classical methods in SR and their contributions in the light of past studies

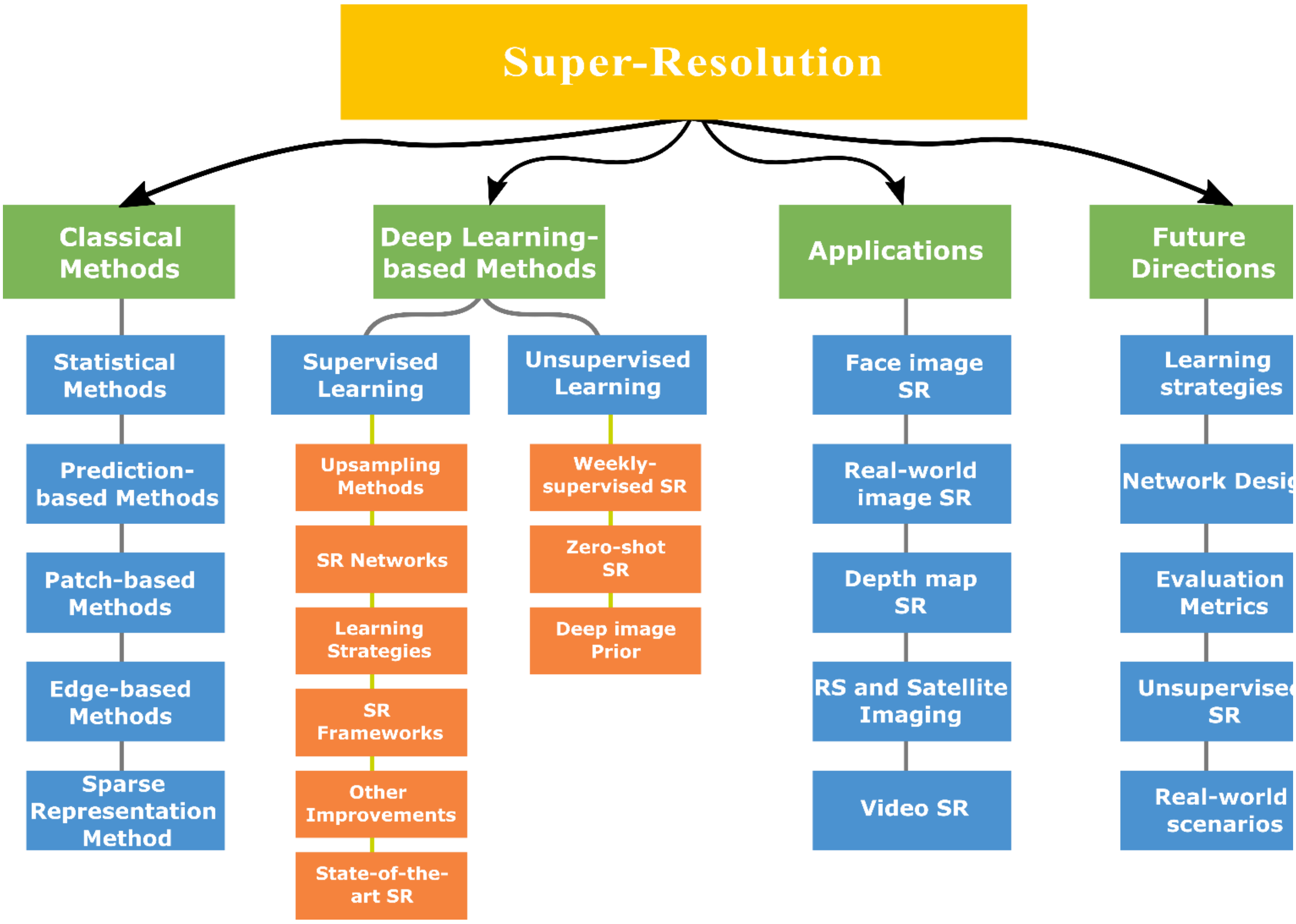

Fig. 2. Hierarchical classification of this survey – Four main categories are (a) classical methods of image super-resolution (b) deep learning-based methods for SR, (c) applications of super-resolution (d) future research and directions in SR

to give perspective.

2. We provide a detailed survey of deep learning-based SR, which includes the definition of the problem, dataset details, performance evaluation, deep learning methods used for SR, and specific applications where these SR methods were used and their performance.

3. We compare the recent advances in the field of deep learning-based SR methods by summarizing the bounds of the methods by providing details of components of the SR methods used structurally.

4. Finally, the open problems in SR and critical challenges that require further probing are highlighted in this survey to provide future directions in the field of SR.

The sections in this study are organized in the following way:

In Section 1, we have introduced the concept of SR and the problem definition and details of the evaluation dataset. In Fig. 2, we have summarized the hierarchical structure of this review. In Section 2, we compare the classical methods of SR, whereas, in Section 3, the SR methods based on supervised deep learning are explored. Section 4 covers the studies that used unsupervised deep learning-based methods for SR. In Section 5, various field-specific applications of SR in recent years are discussed. Finally, in Section 6, we give an overview of the open challenges and limitations in current SR methods and put forward future research directions.

# 2 CONVENTIONAL METHODS OF SUPER-RESOLUTION

Classical methods of SR are briefly discussed in this section to encompass the overall development cycle of the SR. The classical methods include prediction-based, edge-based, statistical, patch-based, and sparse representation methods.

The primary methods were based on prediction, and the first method [85] was based on Lanczos filtering, which filtered the digital data using sigma factors (with modifiable weight function), and similar frequency-domain filtering approach was used in [3] for image resampling. In contrast, cubic convolution [86] was used for resampling the image data, and the results showed that this prediction method was more accurate than the nearest-neighbor prediction algorithm and linear interpolation of image data [87]. In [3], the authors did not consider the blur in the imaging process, while [88] used the knowledge of the imaging process and the relative displacements for image interpolation when the sampling rate was kept constant and this method reduced to deblurring.

The patch-based approach was used in [1], the authors used a training set where various patches within the training set were extracted as training patterns, which helped generate detailed high-frequency images using the patch texture information. Whereas in [89], the authors used locally linear embedding to use local patches for generating high-resolution images based on the local patch features. In contrast, [90] used the concept of reoccurrence of geometrically similar patches in natural images to select the best possible pixel value based on the patch redundancy on the same scales. In [91], the authors introduced the concept of hallucination, where they extracted local features within the LR image first and used these to map the HR image.

Edge-based methods use edge smoothness priors to upsample images, and in [92], a generic image prior, gradient prior profile was used to smoothen the edges within an image to achieve super-resolution in natural images. In [93], the authors used specially designed filters to search for similar patches using the local self-similarity observation, which performed lower nearest patch computations; this method was able to reconstruct realistic-looking edges whereas it performed poorly in clustered regions with fine details.

Statistical methods were used to perform image super-resolution as in [94], where the authors used Kernel ridge regression (KRR) with gradient descent to learn the mapping function from the image example pairs. Adaptive regularization was used to supervise the energy change during the image resampling iterative process; this provided more accurate results as the energy map was used to limit the energy change per iteration, which reduced the noise while maintaining the perceptual quality [95] while [71], [96] used sparse representation methods to perform image super-resolution which used the concept of compressed sensing.

Robust SR method proposed in [97] used the information of outliers to improve the performance of SR in patches where other methods introduce noise due to these outliers. Additionally, [98] proposed a post-processing model that enhanced the resolution of a set of images using a single reference image up to 100x scaling factor. Another way to achieve SR is to use a set of LR images to achieve a single HR image as in [99]. The conventional upsampling methods such as interpolation-based use the information within the LR image to generate HR images, and these methods do not add any new information to the image [100]. Furthermore, they also introduce some inherent problems, such as noise amplification and blur enhancement. Thus, in recent years, the researchers have shifted to learning-based upsampling methods, which are explored in Section 3.

# 3 SUPERVISED SUPER-RESOLUTION

A variety of deep learning methods were developed over the years to solve the SR problem; in this section, the models discussed are trained using both low and high-resolution images (LR-HR pairs). Although there are significant differences in the supervised SR models, and the models can be classified based on the components like the upsampling method employed, deep learning network, learning algorithm, and model frameworks. Any supervised image SR model is based on the combinations of these components, and in this section, we summarize the employed methods for these four components in light of recent supervised image SR research studies.

The component-based review of various methods is performed in this section, and the basic overview of the models is shown in Fig. 2.

## 3.1 UPSAMPLING METHODS

The upsampling is essential in deep learning-based SR methods such as its positioning, and the method performed for upsampling has a significant impact on the training and test performance of the model. There are some commonly used methods [71], [96], [101,102], which use the conventional CNNs for end-to-end learning. In this subsection, various deep learning-based upsampling layers are discussed.

As mentioned in Section 2, the interpolation-based methods of upsampling do not add any new information; hence, in the last decade, learning-based methods are used in image SR.

### 3.1.1 SUB-PIXEL LAYER

The end-to-end learning layer [23] called the sub-pixel layer performs upsampling by generating several additional channels using convolution, and by reshaping these channels, this layer performs upsampling, as shown in Fig. 3. In this layer, convolution is applied to $s_f^2$ where $s_f$ is the scaling factor, as shown in Fig. 3b. Since the input image size is $h \times w \times c$ where $h$ is height, $w$ is width, and $c$ depict color channels, the resulting convolution is $h \times w \times cs_f^2$. In order to achieve the final image, reshuffling [23] operation is performed to get the final output image $s_f h \times s_f w \times c$, as shown in Fig. 3c. Since it is an end-to-end layer, this layer is frequently used in SR models [17], [26], [103], [104].

This layer has a respective wide field, which helps learn more contextual information that results in generating realistic details, whereas this layer may generate some false artifacts at the boundaries of complex patterns due to its uneven distribution of the respective field. Furthermore, predicting the neighborhood pixels in a block-type region sometimes results in unsmooth outputs that do not look realistic when compared with the true HR image, to address this issue, PixelTCL [105] was proposed that used the interdependent prediction layer, which used the information of the interlinked pixels during upsampling. The results were smooth and more realistic when compared with the ground truth image.

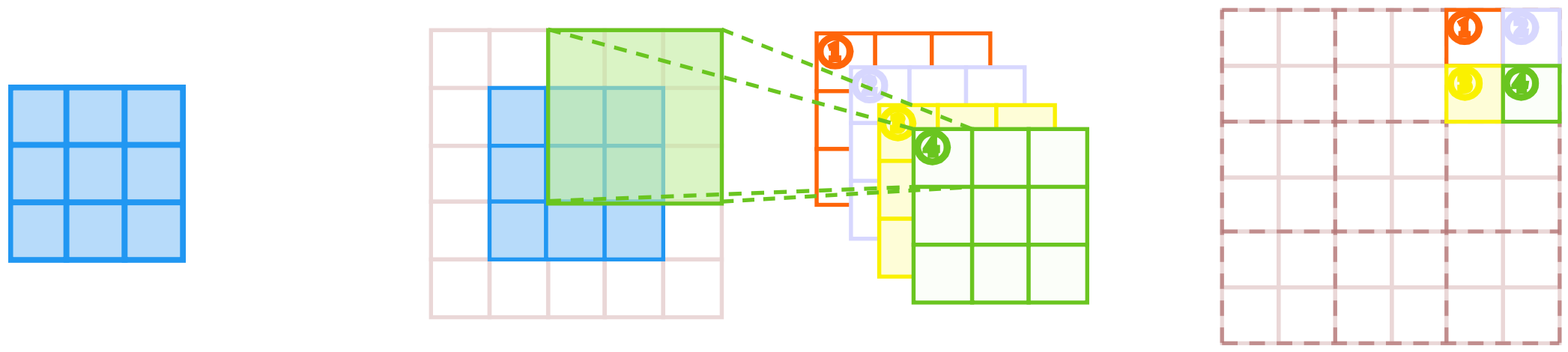

(a) Input (b) Convolution (c) Reshaping

Fig. 3. Sub-pixel layer. Blue color represents the input convolution and output feature maps are represented in other colors

### 3.1.2 DECONVOLUTION LAYER

Deconvolution layer, is also referred to as transposed convolution layer [106], this layer does the converse of the convolution, i.e., predicting the probable input HR-image based on the feature maps from the LR image. In this process, additional zeros are inserted to increase the resolution, and afterward, a convolution is performed. For instance, taking scaling factor 2 for the SR image, a convolution kernel of 3 × 3 (as shown in Fig. 4a, 4b, and 4c), the input LR image is expanded twice by inserting zeros, convolution with the kernel is performed by using a stride and padding of 1.

The deconvolution layer is widely used in SR methods [79], [107–110], as it generates HR images in an end-to-end way, and it has compatibility with the vanilla convolution. As per [111], in some cases, this layer may cause the problem of uneven overlapping within the generated HR image as the patterns are replicated in a check-like format and may result in a non-realistic HR image thereby decreasing the performance of the SR method.

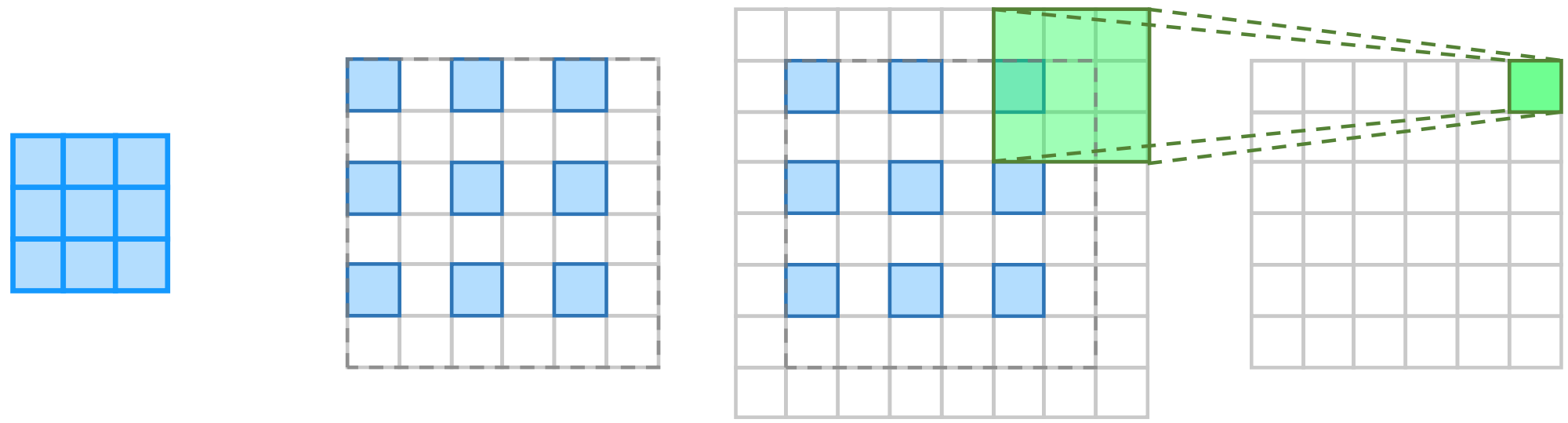

(a) Input (b) Expansion (c) Convolution

Fig. 4. Deconvolution layer. Blue color represents the input, and the green color represents the convolution operation

### 3.1.3 META UPSCALING

In the previously mentioned methods, the scaling factor was predefined, thereby training multiple upsampling modules with different factors, which is often inefficient and is not the actual requirement of an SR method. Meta upscaling module [112] was proposed; this module uses arbitrary scaling factors to generate SR image-based in meta-learning. Meta scaling module projects every position in the required HR image to a small patch in the given LR feature maps ($j \times j \times c_i$), where $j$ is arbitrary, and $c_i$ is the total number of channels within the extracted feature map (in [112] this was 64). Additionally, it also generates the convolution weights ($j \times j \times (c_i \times c_o)$), where $c_o$ is the output image channels, and this is usually 3. Thus, the meta upscaling module continuously uses arbitrary scaling factors within a single model, and using a substantial training set, a large number of factors are simultaneously trained. The performance of this layer even surpasses the results produced with fixed factor models, and even though this module predicts the weights during the inference time, the overall execution time for weight prediction is 100 times less than the total time required for feature extraction [112]. In cases where there is a need for larger magnifications, this module may become unstable as it predicts the convolution weights for every single pixel independent of the image information within those pixels.

In recent years, this method of upscaling is frequently used, particularly in post-upsampling frameworks (Section 3.4.2). The high-level representations extracted from the low-level information are used to construct an HR image using meta upscaling in the last layer of the model, which makes this method an end-to-end SR approach.

## 3.2 DEEP LEARNING SR NETWORKS

The network design and advancements in design architecture are recent trends in deep learning, and in SR, researchers have tried several design implications along with the SR framework (as seen in Section 3.4) for designing the overall SR network. Some of the fundamental and recent network designs are discussed in this section.

### 3.2.1 RECURSIVE LEARNING

One of the basic network-based learning strategies is to use the same module for recursively learning high-level features. This method also minimizes the parameters as the strategy is based on the same module being updated recursively, as shown in Fig. 5a.

One of the most used recursive networks is the Deeply-recursive Convolutional Network (DRCN) [5]. Utilizing a single convolution layer DRCN reaches up to a $41 \times 41$ repetitive field without requiring additional parameters, which is very deep if compared to the Super-resolution Convolution Neural Network SRCNN [22] ($13 \times 13$). The Deep Recursive Residual Network (DRRN) [113] utilized a ResBlock [114] as part of the recursive module for a total of 25 recursions and was reported to achieve better performance than the baseline ResBlock. Using the concept of DRCN, [115] proposed a memory block-based method MemNet which contained six recursive ResBlocks. Whereas the Cascading Residual Network (CARN) [103] also used ResBlocks as recursive units. A recent approach in which the network shares the weights globally in recursion is using an iterative up-and-down sampling-based approach. Apart from end-to-end recursions, the researchers also used Dual-state Recurrent Network (DSRN) [116], which shared the signals between the LR and generated HR states within the network.

Overall, recursive learning networks, while reducing the parameters, can learn the complex representation of the data at the cost of higher computational power. Additionally, the increase in computational requirements may result in an exploding or vanishing gradient. Thus, recursive learning is often used in combination with multi-supervision or residual learning for minimizing the risk of exploding or vanishing gradient [5], [113], [115], [116].

### 3.2.2 RESIDUAL LEARNING

Residual learning was widely used in the field of SR [70], [101], [117], until ResNet [114] was proposed for learning residuals, as shown in Fig. 5b. Overall, there are two approaches, local and global residual learning.

The local residual learning approach mitigates the degradation problem [114], which is caused by increased network depth. Furthermore, the local residual learning also improved the learning rate and reduced the training difficulty; this is frequently used in the SR field [116], [118–120].

The global residual learning is an approach used in which the input and the final output are correlated, and in image SR, the output HR is highly correlated with the input LR image; thus, learning the global residuals between LR and HR image is significant in SR. In global residual learning, the model only learns the residual map that transforms the LR image into an HR image by generating the missing high-frequency details in the LR image. Furthermore, the residuals are minimal, thereby the learning difficulty and model complexity are greatly reduced in global residual-based learning. This method is also frequently used in SR methods [4], [113], [115], [121].

Overall, both methods use residuals to connect the input image with the output HR image; in the case of global residual learning, the connection is directly made, which in local residual learning various layers of different depth to connect the input (using local residuals) with the output.

### 3.2.3 DENSE CONNECTION-BASED LEARNING

Using dense blocks to address the problem of SR, like DenseNet [122], is also a network approach. The dense block utilizes all the features maps generated by the previous layers as inputs along with its feature inputs, which leads to $l(l-1)/2$ connections in an l-layer ($l \geq 2$) dense block. Using dense block, will increase the reusability of the features while resolving the gradient vanishing problem. Furthermore, the dense connections also minimize the model size by utilizing a small growth rate and by enfolding the channels using concatenated input features.

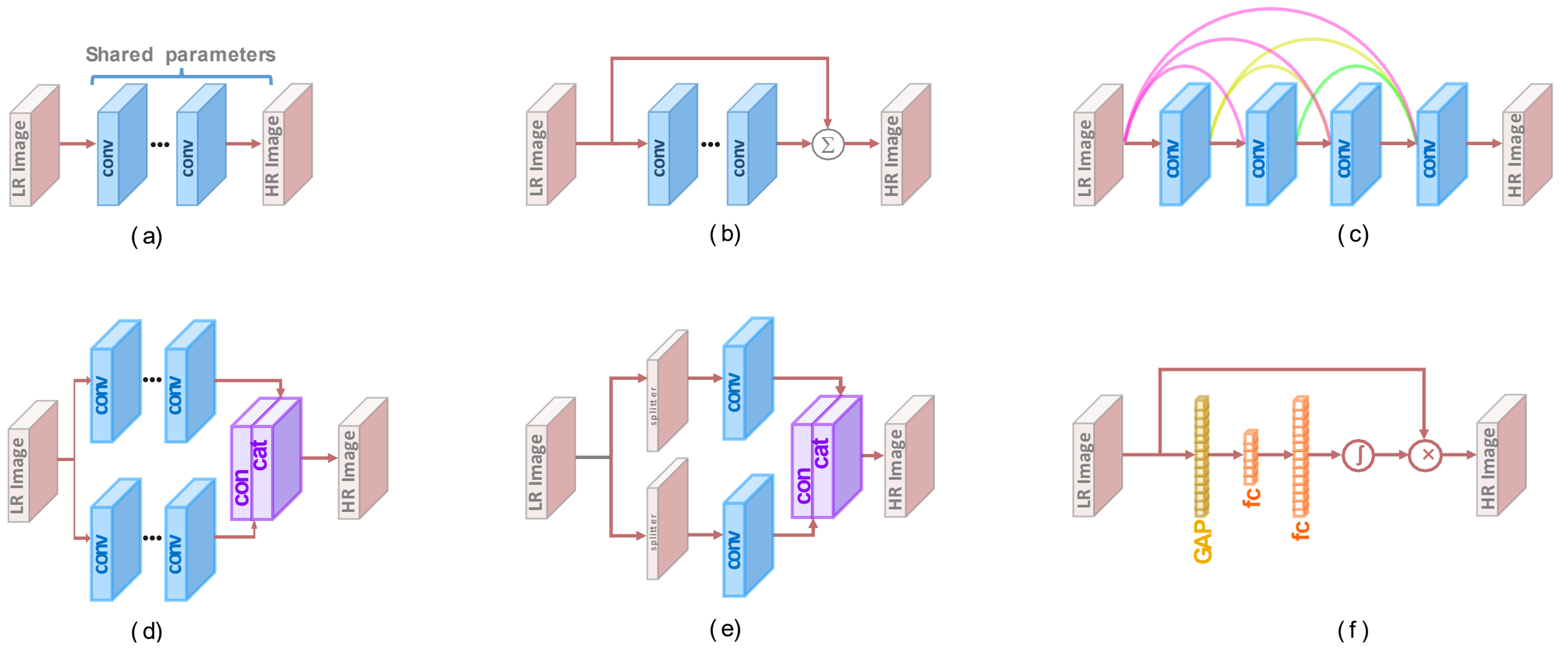

Fig. 5. Deep-learning Network structure for super-resolution. (a) recursive learning, (b) residual learning, (c) dense connection-based learning, (d) multiscale learning, (e) advanced convolution-based learning, (f) attention-based learning

Dense connections are used in SR to connect the low-level and high-level features maps for reconstructing a high-quality fine-detailed HR image, as shown in Fig. 5c. SRDenseNet [79] proposed a 69-layer network that contained dense connections within the dense blocks but also dense connecting among the dense blocks. In SRDenseNet, the feature maps from the prior blocks and the feature maps were used as inputs of all preceding blocks. RDN [104], CARN [103], MemNet [115], and ESRGAN [123] also used layer or block-level dense connection, while DBPN [57] only used the dense connection between the upsampling and downsampling units.

### 3.2.4 MULTI-PATH LEARNING

In multi-path learning, the features are transferred to multiple paths for different representations, and these representations are later combined to gain improved performance. Scale-specific, local, and global multi-path learnings are the main types.

For different scales, the super-resolution models use different feature extraction; in [124], the authors proposed a single network-based multi-path learning for multiple scales. The intermediate layers of the model were shared for feature extraction, while scale-specific paths, including pre-processing and upsampling, were at the end of the models, i.e., the start and end of the network. During training, the scale relative paths are enabled and updated accordingly, and the proposed deep super-resolution MDSR method [124] also decreases the overall model size because of the sharing of parameters across the scales. Like MDSR, a similar multi-path-based approach is also implemented in ProSR and CARN.

Local multi-path learning is inspired using a new block, the inception module [125], for multi-scale feature extraction as performed in MSRN [118] (shown in Fig. 5d). The additional block consists of $3 \times 3$ and $5 \times 5$ kernel size convolution layers, which simultaneously extracts the features. After combining the outputs of the two convolution layers, the final output goes through a $1 \times 1$ kernel convolution. Furthermore, a path links the input and output by element-wise addition, and using this local multi-path learning; this method extracts features in an efficient way than multi-scale learning.

Another variation of multi-path learning is global multi-path learning; in this method, various features are extracted from multi-paths that can interact with each other. In DSRN [116], there are two paths for extracting low and high-level information, and there is a continuous sharing of features for improved learning. In contrast, in pixel recursive SR [126], a conditioning path is responsible for extracting global structures, and the prior path further finds the serial codependence among the generated pixels. A different method was employed by [127], where multi-path learning was performed for unbalanced structures, which were later combined in the final layer to get the SR output.

### 3.2.5 ADVANCED CONVOLUTION-BASED LEARNING

In SR, the basis of the methods explored depend on the convolution operation, and various research studies have attempted to modify the convolution operation for better performance. In recent years, research studies have shown that group convolution, as shown in Fig. 5e, decreased the total number of parameters at the cost of small loops in performance [121], [128]. In CARN-M [103] and IDN [121], group convolution was used instead of the usual vanilla convolution. In dilated convolution, the contextual information is used to generate realistic-looking SR images as in [129], dilated convolution was used to double the receptive field, which resulted in better results.

Another type of convolution is depthwise separable convolution [130], although this convolution significantly reduces the total number of parameters, it reduces the overall performance.

#### 3.2.6 ATTENTION-BASED LEARNING

In deep learning, attention learning is the idea where certain factors are given more preference, which processes the data than others; here, two types of attention-based learning mechanisms are discussed in SR. In channel attention, a particular block is added in the model where global average pooling (GAP) squeezes the input channels; these constants are processed by two fully connected layers to generate channel-wise residuals [131] as shown in Fig. 5f. This technique has been incorporated in SR known as RCAN [132], which has improved performance. Instead of GAP, [133] used second-order channel attention (SPCA) module which used second-order feature metric for extracting more data representation using channel-based attention

In SR, most of the models use local fields for the generation of SR pixels, while in a few cases, some textures or patches which are far apart are necessary for generating accurate local patches. In [134], local and non-local attention blocks were used to extract both local and non-local representations between pixel data. Similarly, the non-local attention technique was incorporated by [133] to capture contextual information using a non-local attention method.

#### 3.2.7 WAVELET TRANSFORM-BASED LEARNING

Wavelet transform (WT) [135,136], is a technique of representing textures using high-frequency sub-bands and global structural information in low-frequency sub-bands in a highly efficient way. WT was used in SR to generate the residuals of the HR sub-bands using the sub-bands of the interpolated LR wavelet. Using the WT, the LR image is decomposed, while the inverse WT provides the reconstruction of the HR image in SR. Other examples of WT based SR are Wavelet-based residual attention network (WRAN) [137], multi-level wavelet CNN (MWCNN) [138], and [139], these approaches used a hybrid approach by combining WT with other learning methods to improve the overall performance.

#### 3.2.8 REGION-RECURSIVE-BASED LEARNING

In SR, most of the methods follow the underlying assumption that it is a pixel-independent process; thus, there is no priority to the interdependence among the generated pixels. By using the concept of PixelCNN [140], [126] proposed a method for pixel recursive learning, which performed SR by pixel-by-pixel generation using two networks. The two networks in [126] captured information about pixel dependence and global contextual information within the pixel recursive SR method. Using the mean opinion scoring evaluation method [126] performed well in comparison to other methods for generating SR face images using the pixel recursive method. The attention-based face hallucination method [141] also utilized the concept of a path-based attention shifting mechanism to enhance the details in the local patches.

While the region-recursive methods perform marginally better than other methods, the recursive process exponentially increases the training difficulty and computation costs due to long propagation paths.

#### 3.2.9 OTHER METHODS

Other SR networks are also used by researchers such as Desubpixel-based learning [142], xUnit-based learning [143], and Pyramid Pooling-based learning [144].

To improve the computational speed, the desubpixel-based approach was used to extract features in a low-dimensional space, which does the inverse task of the sub-pixel layer. By segmenting the images spatially and using them as separate channels, the desubpixel-based learning avoids any information loss, after learning the data representations in low-dimensional space, the images are upsampled to get a high-resolution image. This technique is particularly efficient in applications with limited resources such as smartphones.

In xUnit learning, a spatial activation function was proposed for learning complicated features and textures. In xUnit, the normally used ReLU operation was replaced by xUnit to generate the weight maps through Gaussian gating and convolution. The model size was decreased by 50% using xUnit at the cost of increased computational demand without compromising the SR performance [143].

### 3.3 LEARNING STRATEGIES

Learning strategies also dictate the overall performance of any SR algorithm as the evaluations are dependent upon the choice of the learning strategy selected. In this section, recent research studies are discussed in light of the use of the learning strategy utilized in SR, and some of the critical strategies are discussed in detail.

### 3.3.1 LOSS FUNCTIONS

For any application in deep learning, the selection of the loss functions is critical, and in SR, these functions are used for measuring the error in the reconstruction of HR, and this further helps in the model for optimization iteratively. Since the necessary element of the images is a pixel, initial research studies employed the pixel loss, L2, but it was evaluated that the pixel loss cannot wholly represent the quality of reconstruction [145]. Thus, in SR, different loss functions such as content loss [128] or adversarial loss [17] are used for measuring the error in the generation, and these loss functions have been widely used in the field of SR. Various loss functions are explored in this section, and the notation follows the previously defined variables except where defined otherwise.

**Content Loss.** The perceptual quality, as mentioned previously, is essential in the evaluation of an SR model, and this loss was used in SR [128], [146] to measure the differences between the generated and ground-truth images using an image classification network (*N*). Let the high-level data representation on the $lth$ layer is $r^l(I)$, the content loss is defined as the Euclidean among the high-level representations of the two images *I* and $\hat{I}$, where *I* is the original image and $\hat{I}$ is the generated SR image as below:

$$\mathcal{L}\left(I,\hat{I};N_c,l\right)=\frac{1}{h_l w_l c_l}\sqrt{\sum_{i,j,k}\left(r^l_{i,j,k}\left(\hat{I}\right)-r^l_{i,j,k}\left(I\right)\right)^2} \tag{15}$$

Where $h_l$, $w_l$, and $c_l$ respectively are height, width, and several channels of the image representations in the $l$ layer.

The purpose of content loss is to share the information about image features from the image classification network $N_c$ to the SR network. This loss function ensures the visual similarity between the original image (*I*) and the generated image ($\hat{I}$) by comparing the content and not the individual pixels. Thus, this loss function helps in producing visually perceptible, and more realistic looking images in the field of SR as in [17], [57], [78], [123], [128], [147] where the networks used as pretrained CNNs were ResNet [114] and VGG [148].

**Adversarial Loss.** In recent years, after the development of GANs [149], GANs have received more consideration due to their ability to learn and self-supervise. A GAN combines dual networks performing generation and discrimination tasks, i.e., generating the actual output and using a discriminator network to evaluate the results of the generative network. While training the GANs, there are two continuous updates performed, i.e. (i) Adjust the generator for better results, while training the discriminator to be able to discriminate more efficiently and (ii) Adjust the discriminator while training the generator. This is a recursive training network, and through many iterations of training and evaluation, the generator can generate the output that conforms to the distribution of the actual data. The discriminator is unable to differentiate between real and generated information.

In terms of image SR, the purpose of a generative network is to generate an HR image, while another discriminator network will be used to evaluate if the image is of the same distribution as the input data. This method was first introduced in SR as SRGAN [17], the adversarial loss in [17] was represented by:

$$\mathcal{L}_{GAN_CE_g}\left(\hat{I};D\right)=-\log D\left(\hat{I}\right) \tag{16}$$

$$\mathcal{L}_{GAN_CE_d}\left(\hat{I},I_s;D\right)=-\left\{\log D\left(I_s\right)+\log\left(1-D\left(\hat{I}\right)\right)\right\} \tag{17}$$

Where $\mathcal{L}_{GAN_CE_g}$ is the adversarial loss function of the generator in the SR model, while $\mathcal{L}_{GAN_CE_d}$ is the adversarial loss function of the discriminator D, which in this case, is a binary classifier. In (17), the randomly sampled ground truth image is denoted by $I_s$. The same loss functions were reported by [78].

Other than binary classification error, the studies [150] and [151] used mean square error for improved training and better results compared to [17], the loss functions are given in (18) and (19):

$$\mathcal{L}_{GAN_LS_g}\left(\hat{I};D\right)=\left(D\left(\hat{I}\right)-1\right)^2 \tag{18}$$

$$\mathcal{L}_{GAN_LS_d}\left(\hat{I},I_s;D\right)=\left\{\left(D\left(\hat{I}\right)\right)^2+\left(D\left(I_s\right)-1\right)^2\right\} \tag{19}$$

Contrary to the loss functions mentioned in (18) and (19), [152] showed that in some cases pixel-level discriminator network generates high-frequency noise, thus, we used another discriminator network to evaluate the first discriminator network for high-frequency representations. Using the two discriminator networks, [152] was able to capture all attributes accurately.

Various opinion scoring systems have been used regressively to test the performance of the SR model that uses adversarial loss and although the SR models attained lower PSNR compared to the pixel-loss based SR but on perceptual quality metrics like opinion scoring these adversarial loss based SR methods scored very high [17], [78]. The use of discriminator as the control network for the generator GANs was able to regenerate some intricate patterns that were very difficult to learn using ordinary deep learning methods. The only drawback of the GANs is their training stability [153–156].

**Pixel Loss.** As evident from the name, this loss function performs a pixel-wise comparison between the reference image and the generated image, and there are two types of comparisons, i.e., an *L1* loss which is also termed as mean absolute error and *L2* loss which is the mean square error (MSE)

$$\mathcal{L}_{PIX_L1}\left(I,\hat{I}\right)=\frac{1}{hwc}\sum_{i,j,k}\left|I_{i,j,k}-\hat{I}_{i,j,k}\right| \tag{20}$$

$$\mathcal{L}_{PIX_L2}\left(I,\hat{I}\right)=\frac{1}{hwc}\sum_{i,j,k}\left|I_{i,j,k}-\hat{I}_{i,j,k}\right|^2 \tag{21}$$

The *L1* loss in some cases becomes numerically unstable to compute; thus, another variant of the L1 loss called as the Charbonnier loss [80], [157–159] is given by:

$$\mathcal{L}_{PIX_CH}\left(I,\hat{I}\right)=\frac{1}{hwc}\sum_{i,j,k}\sqrt{\left|I_{i,j,k}-\hat{I}_{i,j,k}\right|^2-e^2} \tag{22}$$

Here *e* is a constant which ensures numerical stability.

The pixel loss function ensures that the generated HR image $\hat{I}$ has the same pixel values as of the HR image *I*. Furthermore, the L2 loss used the square of pixel-value errors, thus give more weightage to high-value differences than lower ones, thus this loss function may give either the too variable result (in case of outliers) or give too smooth results (in case of minimal error values), therefore, the L1 loss function is widely used over L2 loss [103], [124], [160]. Furthermore, the PSNR equation is closely related to the definition of L1 loss, and minimizing L1 loss always leads to increased PSNR. Thus, researchers have often used L1 loss for maximizing the PSNR; as mentioned earlier, the pixel loss function does not cater for the perceptual quality or textures, thus this loss function-based SR networks may have less high-frequency details which result in smooth but unrealistic HR images [36], [60].

**Style Reconstruction Loss.** Ideally, the reconstructed HR image should have comparable styles to the actual HR image (colors, textures, gradient, contrast), thus using the research studies [78], [161], style reconstruction loss was used in SR to match the texture details of the reference image with the generated image. The correlation between the feature maps of different channels as given by the Gram matrix [162] $G^{(l)}$. $G_{i,j}^{(l)}$ is the dot product of the features *i* and *j* in the layer *l*, it is which is given by:

$$G_{i,j}^{(l)}=vec\left(ch_i^{(l)}\left(I\right)\right).vec\left(ch_j^{(l)}\left(I\right)\right) \tag{23}$$

Where *vec()* is the vectorization operation and $ch_j^{(l)}$ denoted the *ith* channel of feature maps in the layer *l*. Now the texture loss is given by (24)

$$\mathcal{L}_{TEX}\left(I,\hat{I};ch,l\right)=\frac{1}{c_l^2}\sqrt{\sum_{i,j}\left(G_{i,j}^{(l)}\left(I\right)-G_{i,j}^{(l)}\left(\hat{I}\right)\right)^2} \tag{24}$$

Using the texture loss function in (24), EnhanceNet [78] reported more realistic results that look visually similar to the reference HR image. Although an optimized texture loss function-based SR generates more realistic-looking images, the selection of patch size is still an open field of research. The selection of small patch size leads to the generation of artifacts in the textured region, while selecting a big patch size generates artifacts across the whole image as the patches are averages over the whole image.

**Total Variation Loss.** Using the pixel values of the neighboring pixels, the total variation loss [163] was defined as the sum of the absolute difference among the values of the neighboring pixels as:

$$\mathcal{L}_{TV}\left(\hat{I}\right)=\frac{1}{hwc}\sum_{i,j,k}\sqrt{\left(\hat{I}_{i+1,j,k}-\hat{I}_{i,j,k}\right)^2-\left(\hat{I}_{i,j+1,k}-\hat{I}_{i,j,k}\right)^2} \tag{25}$$

Total variation loss was used in [17], [150], to ensure smoothness across sharp edges/transitions within the generated image.

**Cycle Consistency Loss.** Using the CycleGAN [164] image SR method was presented in [150] using the cyclic consistency loss function. Using the generated HR image $\hat{I}$, the network generated another LR image $I'_{LR}$, which is further compared with the input LR image $I_{LR}$ for cyclic consistency.

In practice, various loss functions are used as a combination in SR to ensure various aspects of the generation process in the form of a weighted average as in [4], [57], [78], [80]. The selection of appropriate weights of the loss functions in itself is another learning problem as the results vary significantly by varying the weights of the loss function in image SR.

### 3.3.2 CURRICULUM LEARNING

In the Curriculum learning technique [165], the method adapts itself to the variable difficulty of tasks, i.e., starting from simple images with minimum noise to complex images. Since SR always suffers from the adverse condition, the curriculum approach is mainly applied to its learning difficulty and network size. For reducing the training difficulty of the network in SR, small scaling factor, SR is performed in the beginning; in the curriculum learning-based SR, the training starts with 2x upsampling and gradually the next scaling factors 4x, 8x, and so on are generated using the output of previously trained networks. ProSR [151] uses the upsampled output of the previous level and linearly trains the next level using the previous one, while ADRSR [166] concatenates the HR output of the previous levels and further adds another convolution layer. In CARN [167], the previously generated image is entirely replaced by the next level generated image, thus updating the HR image in sequential order.

Another alternative is to transform the image SR problem into N subsets and gradually solving these problems, as in [168], the 8x upsampling problem was divided into three problems (i.e., 1x – 2x; 2x – 4x and 4x – 8x) and three separate networks were used to solve these problems. Using a combination of the previous reconstruction, the next level was finetuned in this method. The same concept was used in [169] to train the network from low image degradations to high image degradations, thus gradually increasing the noise in the LR input image. Curriculum learning reduces the training difficulty; hence, the total computational time is also reduced.

### 3.3.3 BATCH NORMALIZATION

Batch normalization (BN) was proposed by [170] to stabilize and accelerate the deep CNNs by reducing the internal covariate shift of the network. Every mini-batch was normalized, and two additional parameters were used per channel to preserve the representation ability. Batch normalization is responsible for working on the intermediate feature maps; thus, it resolves the vanishing gradient issue while allowing high learning rates. This technique is widely used in SR models such as [17], [26], [113], [115], [138], [171]. In contrast, [124] claimed that batch normalization-based networks lose the scale information of the generated images; thus, there is a lack of flexibility in the network. Hence, [124] used after the removal of batch normalization used the additional memory to design a large model that had a superior performance compared to BN based network. Other studies [123], [151], [172] also implemented this technique to achieve marginally better performance.

### 3.3.4 MULTI-SUPERVISION

Using numerous supervision signals within the same model for improving the gradient propagation and evading the exploding/vanishing gradient problem is called multi-supervision. In [5], multi-supervision is incorporated within the recursive units to address the gradient problems. In SR, the multi-supervision learning technique is implemented by catering a few other factors in the loss function, which improves the back-propagation path and reduces the training difficulty of the model.

## 3.4 SR FRAMEWORKS

SR being an ill-posed problem; thus, the process of upsampling is critical in defining the performance of the SR method. Based on learning strategies, upsampling methods, and network types, there are several frameworks for SR, here four of them are discussed in detail, especially in light of the upsampling method used within the framework, as shown in Fig. 6 – 9.

### 3.4.1 PRE-UPSAMPLING SR

Learning the mapping functions for upsampling from an LR image directly to an HR image is done using this framework, where the LR image is upsampled in the beginning, and various convolution layers are used to extract representations in an iterative way using deep neural networks. Using this concept [16], [63] introduced the pre-upsampling-based SR framework (SRCNN), as shown in Fig. 6. SRCNN was used to learn the end-to-end mapping of LR-HR image conversion using CNNs. Using the classical methods of upsampling as discussed in Section 2, the LR image is firstly converted to HR image, and then deep CNNs were used to learn the representations for mapping the HR image.

Since the pre-upsampling layer already performs the actual pixel conversion task, the network needs to refine the results using CNNs; this results in reduced learning difficulty. Compared to single-scale SR [4], which uses specific scales of input, these models are capable of handling any random size image for refinement and have similar performance. In recent years, many application-oriented research studies have used this framework [5], [25], [113], [115], the differences in these models are in the deep learning layers employed after the upsampling. The only drawback in this model is the use of a predefined classical method of pre-upsampling, which often results in the introduction of image blur, noise amplification in the upsampled image, which later affects the quality of the concluding HR image. Moreover, the dimensions of the image are increased at the start of the method. Thus, the computational cost and space memory requirements of this framework are higher than in other frameworks [23].

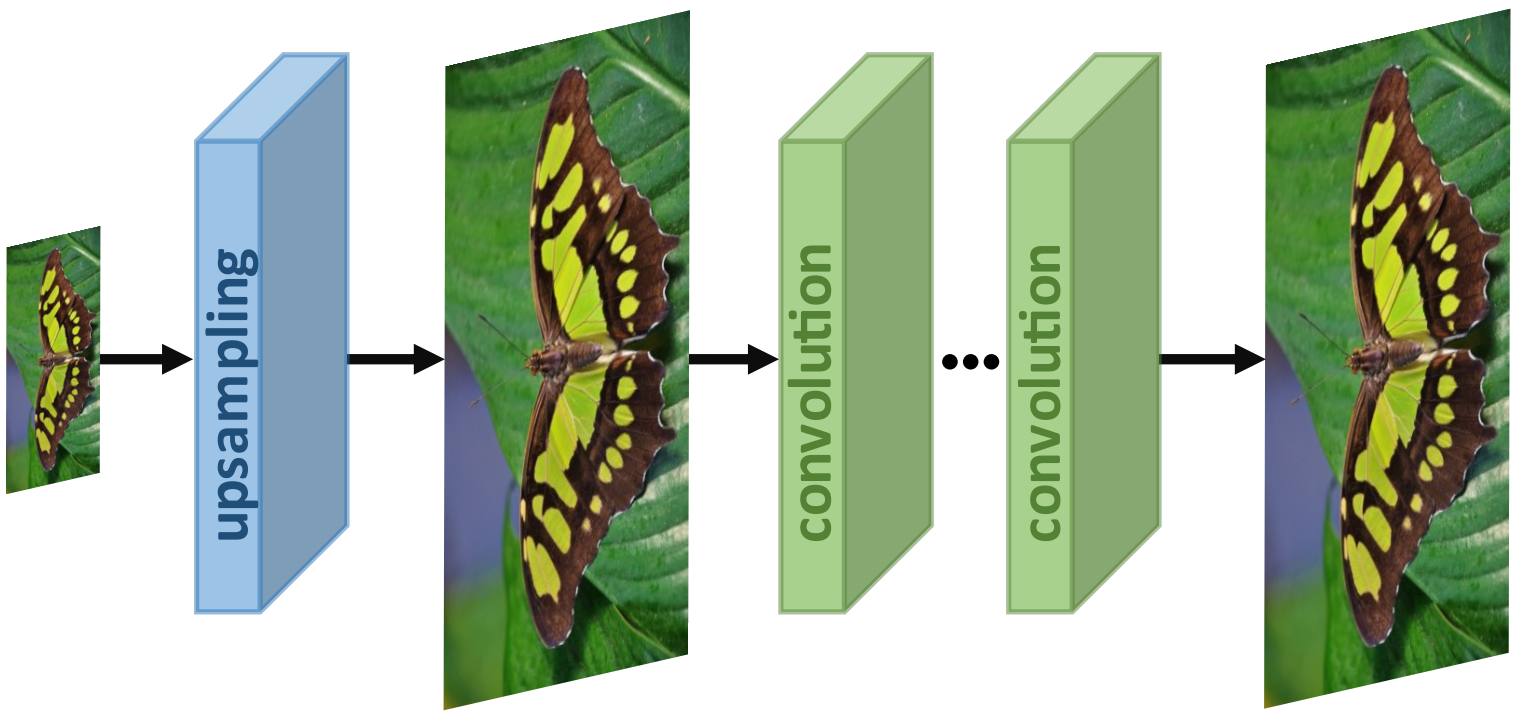

Fig. 6. Pre-upsampling-based super-resolution network pipeline

### 3.4.2 POST-UPSAMPLING SR

To minimize the memory requirements and to increase the computational efficiency, the post-upsampling method was used in SR to utilize deep learning to learn the mapping functions in low-dimensional space. This concept was first used in SR by [23] and [74], and the network diagram is shown in Fig. 7.

Due to low computational costs and the use of low-dimensional space for deep learning this model has been widely used in SR because this reduces the complexity of the model [17], [79], [116], [124].

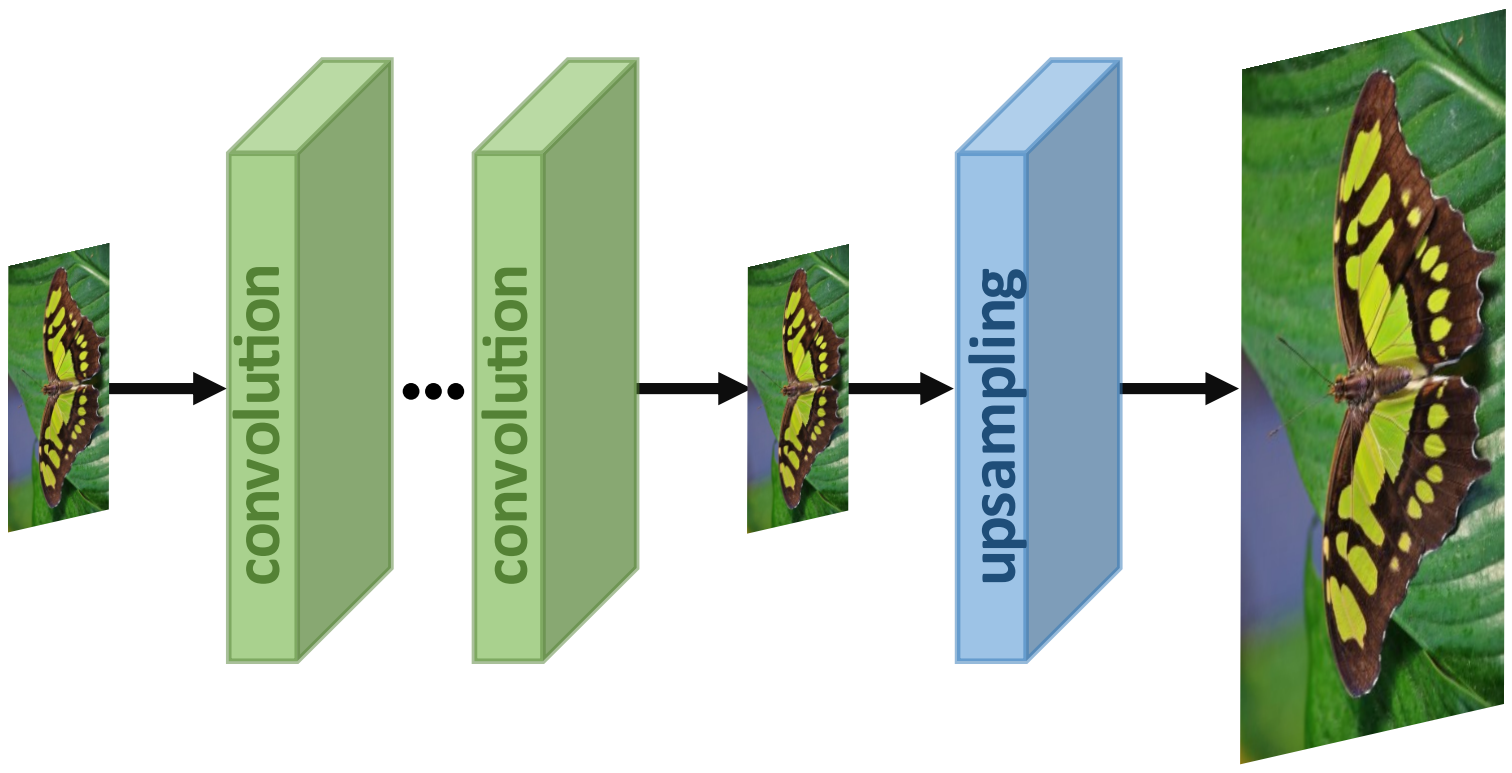

Fig. 7. Post-upsampling-based super-resolution network pipeline

### 3.4.3 ITERATIVE UP-AND-DOWN SAMPLING SR

Since the LR-HR mapping is an ill-posed problem, thus efficient learning using the LR-HR image pair using back-propagation [88] was used in SR by [75]. The SR network is called the iterative up-down sampling SR, as shown in Fig. 8. This model refines the image using recursive back-propagation, i.e., continuously measuring the error and refining the model based on the reconstruction error. The DBPN method proposed in [110], used this concept to perform continuous upsampling and downsampling, and the final image was constructed using the intermediate generations of the HR image.

Similarly, SRFBN [169] used this technique with densely connected layers for image SR, while RBPN [173] used recurrent back-propagation with iterative up-down upsampling for video SR. This framework has shown significant improvement over the other frameworks; still, the back-propagation modules and their appropriate use require further exploration as this concept is recently introduced.

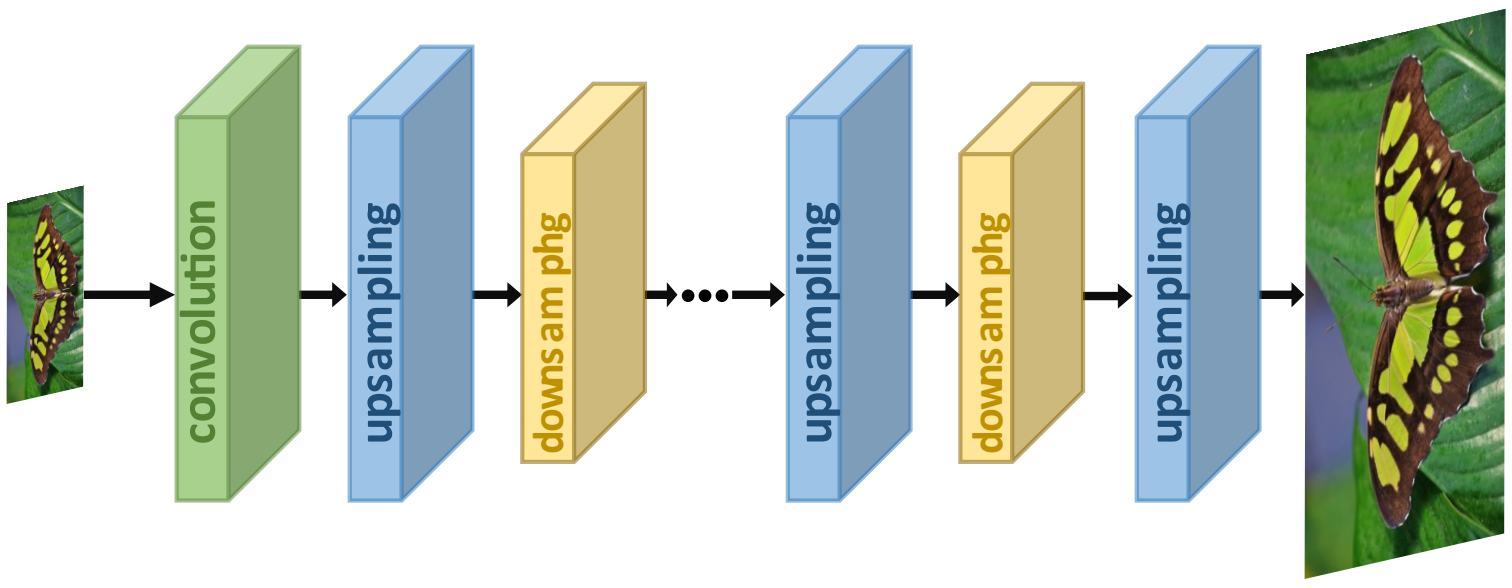

Fig. 8. Iterative up-and-down sampling-based super-resolution network pipeline

### 3.4.4 PROGRESSIVE-UPSAMPLING SR

Since the post-upsampling framework uses a single layer at the end of upsampling and the learning is fixed for scaling factors; thus, multi-scale SR will increase the computational cost of the post-upsampling framework. Thus, using progressive upsampling within the framework to gradually achieve the required scaling was proposed, as seen in Fig. 9. An example of this framework is the LapSRN [80], which uses cascaded CNN based modules that are responsible for the mapping of a single scaling factor, and the output of one module acts as the input LR image to the other module. This framework was also used in ProSR [151] and MS-LapSRN [80].

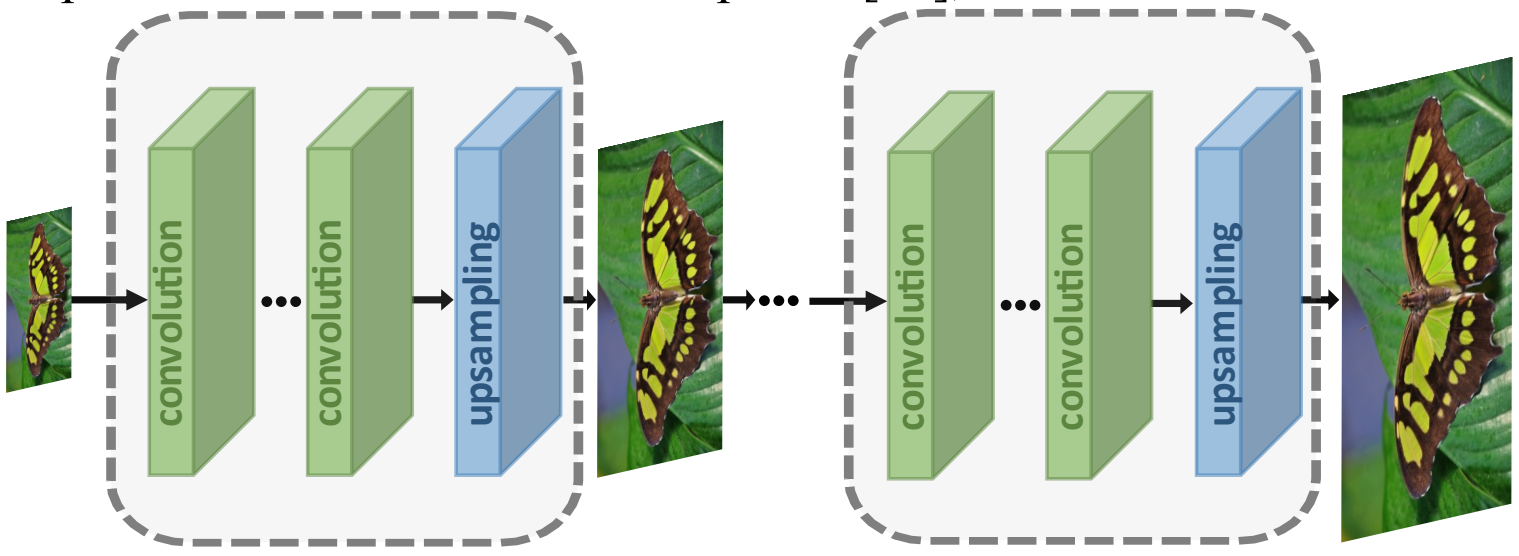

Fig. 9. Progressive sampling-based super-resolution network pipeline

This model achieves higher learning rates as the learning difficulty is less since the SR operation is segregated into several small upscaling tasks, which is more straightforward for CNNs to learn. Furthermore, this model has built-in support for multi-scale SR as the images are scaled with various intermediate scaling factors. Training stability and convergence are the main issues with this framework, and this requires further research.

## 3.5 OTHER IMPROVEMENTS

Apart from the four primary considerations in image SR, other factors have a significant effect on the performance of a super-resolution method, and in this section, a few are discussed in light of recent research.

### 3.5.1 DATA AUGMENTATION

The concept of data augmentation is common in the realm of deep learning, and this concept is used to further enhance the performance of a deep learning model by generating more training data using the same dataset. In the case of image super-resolution, some of the augmentation techniques are flipping, cropping, angular rotation, skew, and color degradation [75], [80], [113], [116], [124]. Recoloring the image using channel shuffling in the LR-HR image pair is also a technique used as data augmentation in image SR [166].

### 3.5.2 ENHANCED PREDICTION

This method of data augmentation affects the output HR image as multiple LR images are augmented using rotation and flipping functions [75]. These augmented are fed to the model for reconstruction, the reconstructed outputs are inversely transformed, and the final HR image is based on the mean [75], [151] or median [25] pixel values of the corresponding augmented outputs.

TABLE 2
SR METHOD DETAILS OF VARIOUS SR ALGORITHMS

| Year | Method Name | US | Network | Framework | Loss Function | Details |
|---|---|---|---|---|---|---|
| 2014, ECCV | SRCNN [16] | Bicubic | CNN | Pre | $\mathcal{L}_{L2}$ | First deep learning-based SR |
| 2016, CVPR | DRCN [5] | Bicubic | Res., Rec. | Pre | $\mathcal{L}_{L2}$ | Recursive layers |
| 2016, ECCV | FSRCNN [74] | Deconv | | Post | $\mathcal{L}_{L2}$ | Lightweight |
| 2017, CVPR | ESPCN [174] | Sub-pixel | | Pre | $\mathcal{L}_{L2}$ | Sub-pixel |
| 2017, CVPR | LapSRN [80] | Bicubic | Res. | Prog | $\mathcal{L}_{L1}$, $\mathcal{L}_{PIX_CH}$ | Cascaded CNN |
| 2017, CVPR | DRRN [113] | Bicubic | Res., Rec. | Pre | $\mathcal{L}_{L2}$ | Recursive layer blocks |
| 2017, CVPR | SRResNet [17] | Sub-pixel | Res. | Post | $\mathcal{L}_{L2}$ | Content loss |
| 2017, CVPR | SRGAN [17] | Sub-pixel | Res. | Post | $\mathcal{L}_{GAN}$ | GAN-based loss |
| 2017, CVPR | EDSR [124] | Sub-pixel | Res. | Post | $\mathcal{L}_{L1}$ | Compact design |
| 2017, ICCV | EnhanceNet [78] | Bicubic | Res. | Pre | $\mathcal{L}_{GAN}$ | GAN-based loss |
| 2017, ICCV | MemNet [115] | Bicubic | Res., Rec., Dense | Pre | $\mathcal{L}_{L2}$ | Memory layers blocks |
| 2017, ICCV | SRDenseNet [79] | Deconv | Res., Dense | Post | $\mathcal{L}_{L2}$ | Fully connected layers |
| 2018, CVPR | DBPN [110] | Deconv | Res., Dense | Iter | $\mathcal{L}_{L2}$ | Back-prop. Based |
| 2018, CVPR | DSRN [116] | Deconv | Res., Rec. | Pre | $\mathcal{L}_{L2}$ | Dual-state network |
| 2018, CVPRW | ProSR, ProGanSR [151] | Progressive Upscale | Res., Dense | Prog | $\mathcal{L}_{LS}$ | Least square loss |
| 2018, ECCV | MSRN [118] | Sub-pixel | Res. | Post | $\mathcal{L}_{L1}$ | Multi-path |
| 2018, ECCV | RCAN [132] | Sub-pixel | Res., Attent. | Post | $\mathcal{L}_{L1}$ | Attention-based loss |
| 2018, ECCV | ESRGAN [123] | Sub-pixel | Res., Dense | Post | $\mathcal{L}_{L1}$ | GAN-based loss |
| 2019, CVPR | Meta-RDN [112] | Meta Upscale | Res., Dense | Post | $\mathcal{L}_{L1}$ | Multi-scale model |
| 2019, CVPR | Meta-SR [112] | Meta Upscale | Res., Dense | Post | $\mathcal{L}_{L1}$ | Arbitrary scale factor as input |
| 2019, CVPR | RBPN [173] | Sub-Pixel | Rec. | Post | $\mathcal{L}_{L1}$ | Used SISR and MISR together for VSR |
| 2019, CVPR | SAN [133] | Sub-Pixel | Res., Attent. | Post | $\mathcal{L}_{L1}$ | 2nd order attention |
| 2019, CVPR | SRFBN [169] | Deconv | Res., Rec., Dense | Post | $\mathcal{L}_{L1}$ | Feedback path |
| 2020, Neuro-computing | WRAN [137] | Bicubic | Res., Attent. | Pre | $\mathcal{L}_{L1}$ | Wavelet-based |

In Table 2, "US," "Rec.," "Res.," "Attent.," "Dense," "Pre.," "Post.," "Iter.," and "Prog." represent upsampling methods, recursive learning, residual learning, attention-based learning, dense connections, pre-upsampling framework, post-upsampling framework, iterative up-down upsampling framework, and progressive upsampling framework respectively.

### 3.5.3 NETWORK FUSION AND INTERPOLATION

This technique used multiple models for prediction of the HR image, and each prediction acts as the input to the next network, like in context-wise network fusion (CNF) [127]. The CNF was based on three individual SRCNNs, and this model achieved the performance, which was in comparison with the state-of-the-art SR models [127].

In the SR network, interpolation is a model thus uses PSNR-based and GAN based models for image SR to boost the performance of SR. Network interpolation strategy [123], [175] used a PSNR-based model for training. In contrast, a GAN-based model was used for fine-tuning while the parameters were interpolated to get the weights of interpolation, and their results had few artifacts and look realistic.

### 3.5.4 MULTI-TASK LEARNING

Multi-task learning is used for learning various problems and getting a generalized model for representations found in learning. For example, image segmentation, object detection, and facial recognition [176,177]. In the field of super-resolution, [57] used semantic maps as input to the model and predicted the parameters of the affine transformation on the transitional feature maps. The SFT-GAN in [57] was able to generate more realistic and crisp looking images with good visual details regarding the textured regions. While in DNSR [166], a denoising network was proposed to denoise the output generated by the SR network; thus, by using this closed-loop system, [166] was able to achieve good results. Like DNSR, [150] proposed an unsupervised SR using the cycle-in-cycle GAN (CinCGAN) for denoising during the SR task. Using a multi-tasking framework may increase the computational difficulty, but the performance of the system increases in terms of PSNR and perceptual quality indexes.

## 3.6 STATE-OF-THE-ART SR METHODS

The recent year has excelled in the development of SR models, especially using supervised deep learning; thus, the models have excelled in achieving state-of-the-art performance. Previously various aspects of the SR models and their underlying components were discussed in light of their strengths and weaknesses. In recent times the use of multiple learning strategies is common, and most of the state-of-the-art methods have used a combination of these strategies.

The first innovation was the use of the dual-branched network (DBCN) [178] to increase the computational efficiency of the single-branched network by using a smaller number of convolutional layers for representation. Furthermore, in RCAN [132], attention-based learning was used in combination with residual learning, L1 pixel loss function, and subpixel upsampling method to achieve the state-of-the-art results in image SR. Furthermore, various models and their reported results and some key factors are summarized in Table 2.

In previous sections, we discussed various strategies and compared and contrasted them while these are important, the performance of any SR algorithm in comparison to the computational cost and parameters is also vital. In Fig. 10, we have graphically shown the performance of SR methods using PSNR metrics in comparison with their size (represented as several parameters) and computational cost (measured by the number of Multi-Adds). The datasets used in measurements are Set14, B100, and Urban 100; the overall PSNR is the average score over the three datasets while the scaling factor for these models was fixed to 2 ×.

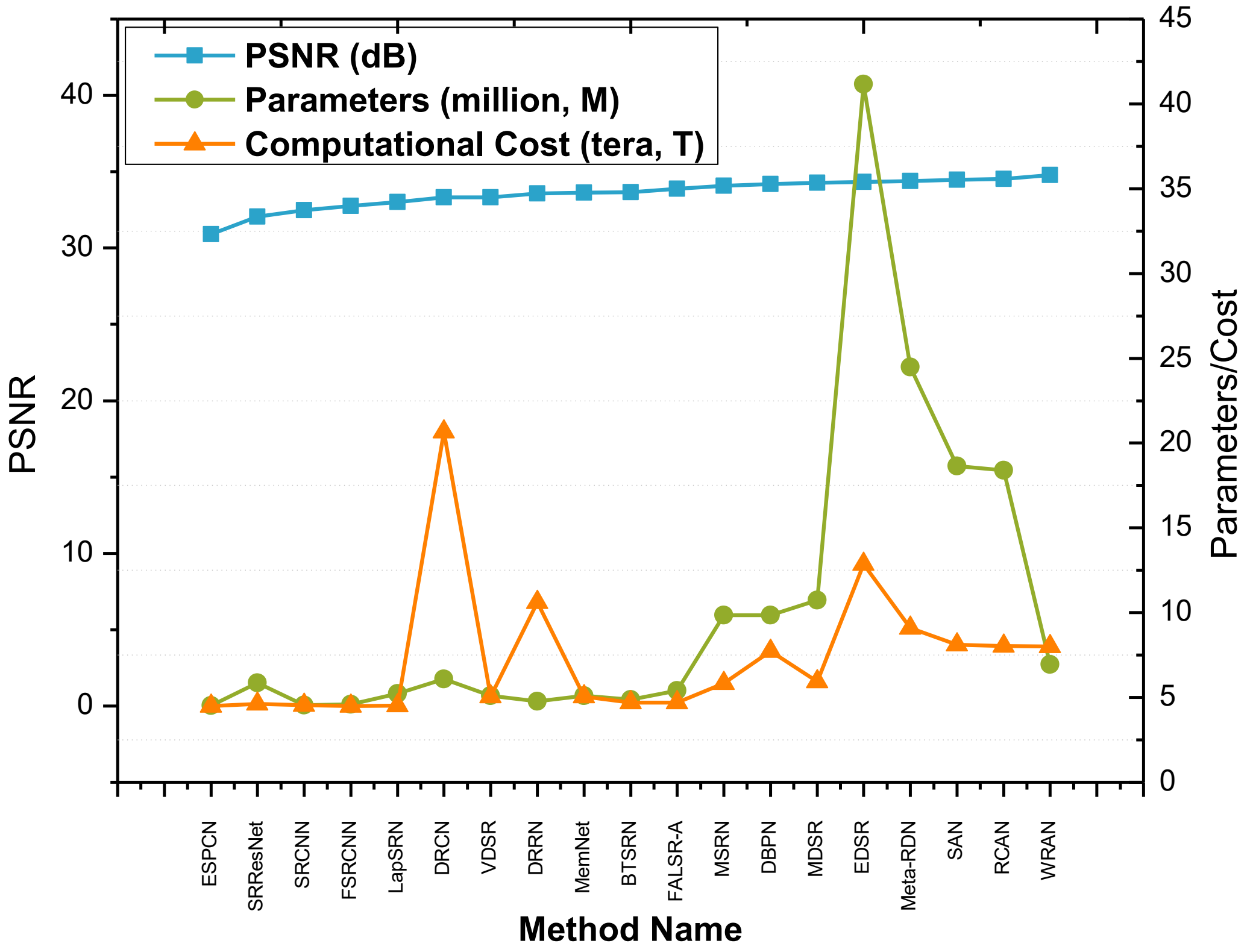

Fig. 10. Benchmarking of super-resolution models. Image quality index is represented by PSNR (in blue color) which is a major evaluation indicator of any super-resolution method, the total number of parameters learnt by every method is shown in green, while the computational efficiency measured by tera multiply-adds are shown in orange color..

# 4 UNSUPERVISED SUPER-RESOLUTION

In this section, the methods of unsupervised SR are discussed, which does not require LR-HR pairs. The limitation of the supervised learning methods is that the LR images are usually generated using known degradations. In supervised learning, the model learns the reverse transformation function of the degradation function to convert the LR image into the HR image. Thus, using the unsupervised model to upsample the LR images is a field of growing interest, where the real-world image degradation is learned by the model to achieve SR using the information of unpaired LR and HR images. A few of the unsupervised SR models are discussed in the sub-sections.

## 4.1 WEAKLY-SUPERVISED SUPER-RESOLUTION

The first method to address the use of known degradation in the model for the generation of LR images with the use of weakly supervised deep learning, this method utilized the unpaired LR and HR images for training the model. Although this model still requires both LR and HR images, the associations are not defined. Thus, there are two possible approaches; the first one is to learn the degradation function first, which can be used to generate the degraded LR images and to train the model to generate the HR images. The other method is to employ degradation function learning and LR-HR mapping cyclically, thus validating the results with each other [179].

### 4.1.1 CYCLIC WEAKLY-SUPERVISED SR

Using the unpaired LR and HR images and referring them to two separated uncorrelated datasets, this method uses a cycle-in-cycle approach to predict the mapping function of these two datasets, i.e., from LR to HR and HR to LR images. This is a recursive process where the mapping functions are tested to generate images with equal distribution and to feed the results to the second prediction in a cyclic way.

Using the deep learning-based CycleGAN [164], a cycle-in-cycle SR framework was proposed in [150], this framework used a total of four generators, while there were two discriminators; the two GANS learned the representation of degraded LR to LR and LR to HR mappings. In [150], the first generator is a simple denoising element that generates similar scale denoised LR images; these denoised images act as input to the second generator to regenerate the HR image, which is further validated by the adversarial network, i.e., a discriminator. Thus, using different loss functions, the CycleGAN achieves image SR using weakly-supervised learning.

Although this method has achieved comparable results, especially in the case of very noisy images where classical degradation functions in supervised learning cannot be used, there is room for research to decrease the learning difficulty of the computational cost of this method.

### 4.1.2 LEARNING THE DEGRADATION FUNCTION

A similar concept to the cyclic SR, but the two networks, i.e., a degradation learning network and LR-HR mapping network, are independently trained. In [180], a two-staged method of image SR was proposed, where a GAN learns the representations of the HR to LR transformation while the second GAN is trained using the paired output of the first GAN to learn the mapping representations of the LR to HR transformation. This two-stage model outperformed the state-of-the-art in Fréchet Inception Distance (FID) [181] with 10% failure cases. This method reported superior reconstruction of HR human facial features.

## 4.2 ZERO-SHOT SUPER-RESOLUTION

Using the concept of training at the time of test, the zero-shot SR (ZSSR) [25] uses a single image to train a deep learning network at the time of test using image augmentation techniques to learn the degradation function. ZSSR was used [182] to predict the degradation kernel, which was further used to generate scaled and augmented images. The final step was to train an SRCNN network to learn the representations of this dataset, and in this way, the ZSSR uses augmentation and input image data to achieve SR. This model outdid the state-of-the-art for non-bicubic, noisy, and blurred LR images by 1dB in case of estimated kernels and 2dB for known kernels.

Since this model requires training for every input image at the time of the test, the overall inference time is substantial.

## 4.3 IMAGE PRIOR IN SR

The low-level details in any learning problem can be mapped using CNNs, thus using a randomly initialized CNN as an image prior was used in [183] to perform SR. The network is not trained; instead, it uses a random vector $v$ as input to the model, and it generates the HR image $I_{yHR}$. The purpose of this method is to determine an image $\hat{I}_{yHR}$ that, when downsampled, returns an LR image that is similar to the input LR image $\hat{I}_{xHR}$. The model performed 2dB the state-of-the-art methods but reported superior results in comparison to the conventional bicubic upsampling method by 1dB.

# 5 DOMAIN-SPECIFIC APPLICATIONS OF SUPER-RESOLUTION

In this section, various applications of SR grouped by the application domains are discussed.

## 5.1 FACE IMAGE SUPER-RESOLUTION

Face hallucination (FH) is perhaps the ultimate target utility of the image SR for face-recognition-based tasks such as [13], [184–187]. The facial images contain facial-structured information; thus, using image priors in FH has been a common approach to achieve FH.

Using techniques such as in CBN [188], the generated HR images can be constrained to face-related features, thereby forcing the model to output HR images that contain facial features. In CBN, this was achieved by using a facial prior and a dense correspondence field estimation. While in FSRNet [40] facial parsing maps and facial landmark heatmaps were used as priors to the learning network to achieve face image SR, SICNN [186] used a joint training approach to recover the real identity using a super-identity loss function. Super-FAN [147] approached FH using end-to-end learning with FAN to ensure the generated images are consistent with human facial features.

Using implicit methods for solving the face misalignment problem is another way to approach FH; for instance, in [189], the spatial transformation is achieved by using transformation networks [190]. Another method based on [190] is TDAE [175], which uses a three-module approach for FH, using a D-E-D (decoder-encoder-decoder) model to achieve FH; the first decoder performs denoising and upsampling while encoder downsamples the denoised image which is fed to the final decoder for FH. Another approach is to use HR exemplars from datasets to decompose the facial features of an LR image and project the HR features into the exemplar dataset to achieve FH [191]. In [192], an adversarial discriminative network was proposed for feature learning on both feature space and raw pixel space; this method performed well for heterogenous face recognition (HFR).

In other research studies, human perception of attention shifting [193] was used in Attention-FH [141] to learn face patches for local enhancement of FH. In [194], a multi-class GAN network was proposed for FH, which composed of multiple generators and discriminators, while in [195], the authors adopted a network model analogous to SRGAN [17]. Using the method of conditional GAN [196], the studies [197,198] used additional facial features to achieve FH with predefined attributes.

## 5.2 REAL-WORLD IMAGE SUPER-RESOLUTION

In the case of real-world images, the sensors used to capture them already introduce degradations as the final RGB (8-bit) image is converted from the raw image (which is usually more than 14-bit or higher). Thus, using these images as reference for SR is not optimal as the images have already been degraded [199]. To approach this problem, research studies such as [200] and [201] have proposed methods for the development of real-world image datasets. In [200], the SR-RAW dataset was developed by the authors which contained raw-HR-LR(RGB) pairs generated using the optical zoom in cameras, while in [201], image resolution and its relationship with the field of view (FoV) were explored by the authors to generate a real-world dataset called City100.

## 5.3 DEPTH MAP SUPER-RESOLUTION

In the field of computer vision, problems like image segmentation [202–204]and pose estimation [205–207] have been approached by using depth maps. Depth maps retain the distance information of the scene and the observer, although these depth maps are of low-resolution because of the hardware constraints of the modern camera systems. Thus, image SR is used in this regard to increase the resolution of the depth maps.

Using multiple cameras to record the same scene and generate multiple HR images is the most suitable way of doing depth map SR. In [208], the authors used two separate CNNs to downsample HR image concurrently and upsample the LR depth map, after the generation of RGB features from the downsampling CNN, these features were used to fine-tune the upsampling process of depth maps. While [209] used the energy minimization model (such as [77]) to guide the model for generating HR depth maps without the need for reference images.

## 5.4 REMOTE SENSING AND SATELLITE IMAGING

The use of SR is enhancing the resolution of remote sensing, and satellite imaging has increased in the past years [210]. In [211], the authors used the concept of multi-line cameras to utilize multiple LR images to generate a high-quality HR image from the ZY-3 (TLC) satellite image dataset. In [212,213], the authors argued that the conventional methods of evaluation of the SR techniques are not valid for satellite imaging as the degradation functions and operation conditions of the satellite hardware are entirely in a different environment and thus [213] proposed a new way for validation of SR methods for satellite image SR methods. An adaptive multi-scale detail enhancement (AMDE-SR) was proposed in [214] to use the multi-scale SR method for the generation of high-detailed HR images with accurate textual and high-frequency information.

### 5.5 VIDEO SUPER-RESOLUTION

In video SR, there are multiple frames for representing the same scene; thus, in the video, there is inter and intra-frame spatial dependency, which includes the information of brightness, colors, and relative motion of objects. Using the optical flow-based method in [215,216] proposed a method to generate probable HR candidate images and ensemble these images using CNNs. Using the Druleas [217] algorithm, CVSRnet [218] addressed the effect of motion by using CNNs for the images in successive frames to generate HR images.

Apart from direct learning motion compensation, a trainable spatial transformer [190], was used in VESPCN [174] to motion compensation mapping using data from successive frames for end-to-end mapping. Using a sub-pixel layer-based module, [6] achieved super-resolution and motion compensation simultaneously.

Another approach is to use recurrent networks to grasp the spatial and temporal interdependency to address the motion compensation indirectly. In STCN [219], the authors used a bidirectional LSTM [220] and deep CNNs to extract the temporal and spatial information from the video frames, while BRCN [221] utilized RNNs, CNNs and conditional CNNs respectively for temporal, spatial and temporal-spatial interdependency mapping. Using 3D convolution filters of small size to replace the large-sized filter, FSTRN [222] achieves state-of-the-art performance by using deep CNNs, sustaining a low computational cost.

### 5.6 OTHER APPLICATIONS

Other fields also used the concept of image super-resolution to achieve high-resolution images, such as in [223], the authors proposed the use of progressive GANs to enhance the image quality of magnetic resonance (MR) images. The DeepResolve [224] used image SR methods to generate thin-sliced knee MR images from the thick-sliced input images. Since the diffusion MRI has high image acquisition time and low resolution, SRR-DTI [225], reconstructed HR diffusion parameters from LR DW images [10] also used SRR-DTI to find the structural sex variances in the adult zebra finch brain.

Other applications of SR include object detection [226,227], stereo image SR [228–230]; overall, SR plays a vital role in multi-disciplines from medical science, computer vision to satellite imaging and remote sensing.

## 6 CONCLUSION AND FUTURE DIRECTIONS

In this survey paper, a detailed survey of classical methods of super-resolution and recent advances in SR with deep learning are explored. The central theme of this survey was to discuss deep learning-based SR techniques, along with the application of SR in various fields. Although image SR has achieved a lot in the last decade, there are still some open problems. In this section, we summarize the future trends in SR. This survey is intended for not only the researchers in the field of SR but also for researchers from other fields to use image SR in their respective fields of interest.

### 6.1 LEARNING STRATEGIES

Learning strategies in image SR are introduced in Section 3.3, while the learning strategies are well matured in image SR, there are research directions in the development of alternate loss functions and alternative of batch normalization

- There are various loss functions in SR, and the choice of SR depends upon the task, while it is still an open research area to find an optimal loss function that fits all SR frameworks. Currently, a combination of loss functions is used to optimize the learning process, and there are no standard criteria for the selection of loss function, thus exploring various probable loss functions for super-resolution is a promising future direction.
- Batch normalization is a technique that performs well in computer vision tasks and reduces the overall runtime of the training and enhances the performance, however, in SR batch normalization proved to be sub-optimal [124], [151], [172]. In this regard, normalization techniques for super-resolution should be explored further.

### 6.2 NETWORK DESIGN

Network design strategies require further exploration in SR as the network design inherently dictates the overall performance of any SR method. Some of the key research areas are highlighted in this section

- Current upsampling methods, as discussed in Section 3.1, have significant drawbacks for the deconvolution layer and may produce checkerboard artifacts. In contrast, the sub-pixel layer is susceptible to non-uniform distribution of receptive fields; the meta-scale method has stability issues, while the interpolation-based methods lack end-to-end learning. Thus, further research is required in the exploration of upsampling methods that can be generic to SR models and can be applied to LR images with any scaling factors.
- For human perception in SR, further research is required in attention-based SR, where the models may be trained to give more attention to some image features than others like the human visual system does.
- Using a combination of low and high-level representations simultaneously to accelerate the SR process is another field in network design for fast and accurate reconstruction of the HR image.
- Exploring network architectures that can be implemented in practical applications since current methods use

deep neural networks, which increases the performance of the SR at the expense of higher computational cost; thus, research in the development of network architecture that is minimal and provides optimal performance is another promising research direction.

### 6.3 EVALUATION METRICS

The image quality metrics used in SR act as the benchmark score, while the two most commonly used metrics, PSNR and SSIM, help gauge the performance of SR, but these metrics introduce inherent issues in the generated image. Using PSNR as an evaluation metric usually introduces non-realistic smooth surfaces, while SSIM works with textures, structures, brightness, and contrast to imitate human perception. Both of these metrics are unable to completely grasp the perceptual quality of images [17], [78]. Opinion scoring is a metric that ensures perceptual quality, but this metric is impractical for implementing SR methods for large datasets; thus, a probable research direction is the development of a universal quality metric for SR.

### 6.4 UNSUPERVISED SUPER-RESOLUTION

In the past two years, unsupervised SR methods have gained popularity, but still, the task of collecting various resolution scenes for a similar pose is difficult; thus, bicubic interpolation is used instead to generate an unpaired SR dataset. In actual, the unsupervised SR methods learn the inverse mapping of this interpolation for the reconstruction of HR images, and the actual learning of SR is still an open research field using unsupervised learning methods.

### 6.5 DOMAIN-SPECIFIC APPLICATIONS

As indicated in Section 5, SR has a promising implementation in other domains for generating HR images, thus using SR in domain-specific applications is another research direction, for instance, in medical imaging techniques and graphics/video rendering.

## ACKNOWLEDGMENT

This work was supported by the Natural Science Basic Research Plan in Shaanxi Province of China (2019JM-311). We show our gratitude to the authors of all referred research studies for sharing results, especially to the authors of [5], [17], [74], [80], [110], [112], [113], [115], [118], [124], [132], [133], [137], [174]. Furthermore, the authors show gratitude to the anonymous reviewers for their helpful comments and suggestions.

## SUPPLEMENTARY MATERIAL

A tabular form of benchmarking results as shown in Figure 10 is appended in supplementary material as Table S1 where the numerical values of all the performance indicators are shown for selected image SR methods.

TABLE 3
LIST OF SYMBOLS

| | |
|---|---|
| $I_{xLR}$ | input low-resolution image |
| $I_{yE}$ | estimated HR corresponding to $I_{xLR}$ |
| $I_{yHR}$ | true high-resolution image |
| $I_r$ | reference image |
| $\partial$ | parameters of the image degradation function |
| $d()$ | image degradation function |
| $g()$ | super-resolution function |
| $\delta$ | parameters to the function $g$ |
| $\downarrow_s$ | downsampling operator |
| $\otimes$ | convolution operator |
| $n_\sigma$ | additive white Gaussian noise with a standard deviation of $\sigma$ |
| $\kappa$ | blue kernel |
| $\Psi(\phi)$ | regularization term |
| $I_y(i)$ | ith pixel value of the image $I_y$ |
| $C_I$ , $L_I$ | contract and Luminance of image $I$ |
| $Com_l$ , $Com_c$ . $Com_s$ | comparison functions of luminance, contrast, and structure |
| $\sigma_{I_r,\hat{I}}$ | covariance between $I_r$ and $\hat{I}$ |
| $r^l(I)$ | high-level data representations in layer $l$ |
| $N_c$ | image classification network |
| $M$ | maximum pixel value |
| $P$ | total number of pixels in an image |
| $\mu_1$ , $\mu_2$ , $\mu_3$ | constants for stability |
| $\alpha$ , $\beta$ , $\gamma$ | control parameters for SSIM function |
| $s_f$ | scale factor |
| $\mathcal{L}$ | loss function |
| $\mathcal{L}_{GAN_CE_g}$ , $\mathcal{L}_{GAN_CE_d}$ | the adversarial loss function of least square-based cross-entropy generator and discriminator |
| $\mathcal{L}_{GAN_LS_g}$ , $\mathcal{L}_{GAN_LS_d}$ | the adversarial loss function of least square-based generator and discriminator |
| $\mathcal{L}_{PIX_L1}$ , $\mathcal{L}_{PIX_L2}$ | L1 and L2 pixel loss |
| $\mathcal{L}_{PIX_CH}$ | L1 Charbonnier loss |
| $G^{(l)}$ | Gram matrix representation |
| $\mathcal{L}_{TEX}$ , $\mathcal{L}_{TV}$ | texture loss, total variation loss |

TABLE 4
LIST OF ACRONYMS

| | |
|---|---|
| AMDE-SR | adaptive multi-scale detail enhancement |
| BN | batch normalization |
| CARN | cascading residual network |
| CinCGAN | cycle-in-cycle GAN |
| CNF | context-wise network fusion |
| CNN | convolutional neural network |
| CVPR | conference on computer vision and pattern recognition |
| DBCN | dual-branched network |
| DBPN | deep back-propagation network |
| D-E-D | decoder-encoder-decoder |
| DenseNet | densely connected network |
| DIP | discriminable image pairs |
| dipQA | DIP inferred quality index |
| DL | deep learning |
| DNSR | denoising for SR |
| DRCN | deeply-recursive convolutional network |
| DRRN | deep recursive residual network |
| DSRN | dual-state recurrent network |
| ECCV | European conference on computer vision |
| EDSR | enhanced deep SR network |
| ESPCN | efficient sub-pixel CNN |
| ESRGAN | enhanced SRGAN |
| FH | face hallucination |
| FID | Fréchet inception distance |
| FoV | field of view |
| FR-IQA | full reference image quality assessment |
| FSIM | feature similarity index metric |
| FSRCNN | fast SRCNN |
| FSRNet | face SR network |
| FSTRN | fast spatio-temporal ResNet |
| GAN | generative adversarial nets |
| GAP | global average pooling |
| HFR | heterogeneous face recognition |
| HR | high-resolution |
| HVS | human visual system |
| IDN | information distillation network |
| IQA | image quality assessment |
| KRR | kernel ridge regression |
| LapSRN | Laplacian pyramid SR network |
| LR | low-resolution |
| MDSR | the multi-scale deep SR system |
| MemNet | memory network |
| MEON | multi-task end-to-end optimized deep neural network |
| MR | magnetic resonance |
| MSE | mean square error |
| MS-LapSRN | multi-scale LapSRN |
| MSRN | multiscale residual network |
| MS-SSIM | multi-scale SSIM |
| MWCNN | multi-level wavelet CNN |
| NTIRE | new trends in image restoration and enhancement |
| PRIM | perceptual image restoration and manipulation |
| ProSR | progressive SR |
| PSNR | peak signal-to-noise ratio |
| RBPN | RBPN |
| RDN | residual dense network |
| ResNet | residual network |
| SAN | second-order Attention Network |
| SFT-GAN | spatial feature transformation GAN |
| SISR | single image super-resolution |
| SPCA | second-order channel attention |
| SR | super-resolution |
| SRCNN | super-resolution convolution neural network |
| SRFBN | SR feedback network |
| SSIM | structural similarity index |
| Super-FAN | FAN-based SR |
| VSRnet | video SR network |
| WRAN | wavelet-based residual attention network |
| WT | wavelet transform |
| ZSSR | zero-shot SR |